%% file: Backdoor.tex
\documentclass[10pt,journal]{IEEEtran}
\IEEEoverridecommandlockouts
\usepackage{cite}
\usepackage{amsmath,amssymb,amsfonts,amsthm}
\usepackage{graphicx}
\usepackage{textcomp}
\usepackage{xcolor}
\usepackage{tcolorbox}
\usepackage{colortbl}

\usepackage[utf8]{inputenc} 
\usepackage[T1]{fontenc}    
\usepackage{hyperref}       
\usepackage{url}            
\usepackage{booktabs}       
\usepackage{nicefrac}       
\usepackage{microtype}      
\usepackage{balance}

\usepackage{xspace}
\usepackage{color}
\usepackage{makecell}
\usepackage{multirow}
\usepackage{pifont}

\usepackage{algorithm}
\usepackage{algpseudocode}

\usepackage{subcaption}
\usepackage{url}
\usepackage{cleveref}

\newcommand{\cmark}{\ding{51}}%
\newcommand{\xmark}{\ding{55}}%

\newtheorem{definition}{Definition}

\newcommand{\revise}[1]{\textcolor{black}{#1}}
\newcommand{\reviseNew}[1]{\textcolor{black}{#1}}
\newcommand{\BD}{\textsc{Neo}\xspace}
\newcommand{\tableBoundaryColor}{white}

\algnewcommand{\LineComment}[1]{\State\(\triangleright\) #1} 
\let\oldReturn\Return
\renewcommand{\Return}{\State\oldReturn}

\def\BibTeX{{\rm B\kern-.05em{\sc i\kern-.025em b}\kern-.08em
    T\kern-.1667em\lower.7ex\hbox{E}\kern-.125emX}}

\author{Sakshi Udeshi,
        Shanshan Peng,  
        Gerald Woo,
        Lionell Loh,
        Louth Rawshan,
        Sudipta Chattopadhyay
\IEEEcompsocitemizethanks{\IEEEcompsocthanksitem S. Udeshi, S. Peng, G. Woo, L. Loh, L. Rawshan 
and S. Chattopadhyay are with the Singapore University of Technology and Design, Singapore.\protect\\
E-mail: sakshi\_udeshi@sutd.edu.sg.}
}

\begin{document}

\title{
Model Agnostic Defence against Backdoor Attacks in Machine Learning
}


\markboth{IEEE Transactions on Reliability}%
{Udeshi \MakeLowercase{\textit{et al.}}: Model Agnostic Defence against Backdoor Attacks in Machine Learning}

\maketitle
\begin{abstract}

Machine Learning (ML) has automated a multitude of our day-to-day 
decision-making domains such as education, employment and driving automation.
The continued success of ML largely depends on our ability to 
trust the model we are using. Recently, a new class of attacks
called Backdoor Attacks have been developed. These attacks undermine
the user's trust in ML models. In this work, we present \BD, a 
model agnostic framework to detect and mitigate such 
backdoor attacks in image classification ML models. For a given 
image classification model, 
our approach analyses the inputs it receives and determines if 
the model is backdoored. In addition to this feature, we also mitigate these
attacks by determining the correct predictions of the poisoned images. 

We have implemented \BD and evaluated it against three state-of-the-art poisoned models. 
In our evaluation, we 
show that \BD can detect $\approx$88\% of the poisoned inputs on average
and it is as fast as 4.4 ms per input image. 
We also compare our \BD approach 
with the state-of-the-art defence methodologies proposed for backdoor attacks. 
Our evaluation reveals that despite being a blackbox approach, \BD is more 
effective in thwarting backdoor attacks than the existing techniques.
Finally, 
we also reconstruct the exact poisoned input for the user to effectively test 
their systems. 
\end{abstract}



\input{introduction}

\input{background}

\input{overview}

\input{methodology}

\input{results}

\input{relatedWork}

\input{threatsToValidity}

\input{conclusion}

\balance
\bibliographystyle{plainurl}
\bibliography{Backdoor}

\end{document}

%% file: introduction.tex
\section{Introduction}
\label{sec:introduction}

Due to the massive progress in Machine Learning (ML) in the last decade, 
its popularity now has reached a variety of application domains, 
including sensitive and safety critical domains, such as automotive, 
finance, education and employment. One of the key reasons to use 
ML is to automate mundane and error prone tasks of 
manual decision making. In light of the broad definition of what 
constitutes an ML system, we restrict our focus to 
systems which are image-based. These systems are trained using 
preliminary images and make decisions based on the new images  
provided to the system. Training such an image-based ML model is 
usually a computationally expensive task and 
it is usually more cost-effective to rely on a third party's 
computing infrastructure and expertise to train these models.
\reviseNew {The users most vulnerable to this type of attack 
are usually users who are not technically sophisticated.}
Thus, it is often 
delegated to the \reviseNew{untrusted third-party service provider 
(e.g. Machine-learning-as-a-service 
providers, individual machine learning consultants) 
who have the required technical knowledge.} Unfortunately, 
this brings along a new attack vector, called {\em backdoor}, 
for ML systems~\cite{BadNets}. The basic idea behind 
a backdoor attack is to poison the training set and train the 
respective algorithm with this poisoned set. The outcome is 
a poisoned image classifier that behaves maliciously 
{\em only for observations that are poisoned}~\cite{BadNets}. 
Backdoor attacks are highly stealthy in nature, as they do not 
reduce the accuracy of the poisoned model on  
clean (i.e. not poisoned) datasets. Thus, these attacks cannot 
be detected by simply comparing the accuracy of the model on a 
pre-defined clean dataset. 
\reviseNew{We also note that such 
service providers might be untrusted not just due to the malicious 
nature, but also due to a compromise from potential attackers, 
for example, due to side channel attacks in the cloud~\cite{crossVM}. 
Such side channel attacks could leak sensitive information about 
the VM which can be exploited to gain access to the VM and the training
process.
}
There is thus a call for efficient 
verification and validation methodologies to detect backdoors 
in ML systems.

Backdoor attacks are launched due to a backdoor trigger embedded 
into the input~\cite{BadNets}\cite{trojannn}. The poisoned model is 
trained to recognise this trigger and the backdoor attack is activated as 
soon as an input with the trigger is presented to the 
model. \reviseNew {To this end, we present \BD, a
model agnostic framework to detect and mitigate such 
backdoor attacks in image classification ML models.}
Given an arbitrary poisoned image presented to the poisoned ML-based 
image classifier, 
it is possible to locate the position of the backdoor trigger 
in the image. Subsequently, \BD covers the trigger to neutralise
the malicious behaviour of the backdoor. This is our main 
intuition. We propose \BD, a novel approach to detect and mitigate backdoor 
attacks for arbitrary ML-based image classifiers. For a given 
classifier, which could be either poisoned or clean, and a stream of 
images given to the classifier, \BD works alongside the classifiers to check 
whether it behaves maliciously due to a backdoor. Moreover, 
\BD can precisely reconstruct the backdoor trigger with which the 
training dataset was poisoned. Thus, \BD not only helps to detect 
and mitigate the effect of backdoor, it also aids the user of the 
classifier by making them aware of the backdoor triggers, 
helping them improve their model. 
By exposing the backdoor trigger, \BD also impairs the stealthy nature of 
this attack.

\begin{figure}[h]
\begin{center}
\includegraphics[scale=0.8]{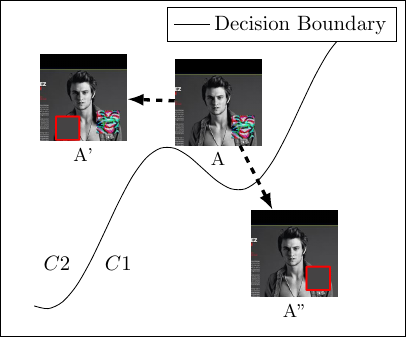}
\end{center}
\caption{The intuition behind \BD (Note that the red outline is only to highlight the trigger blocker)}
\label{fig:neo-idea}
\end{figure}

As an example, consider the decision boundary of a backdoored classifier 
shown in \Cref{fig:neo-idea}. The image {\tt A} is poisoned with the 
backdoor trigger located at the bottom right corner. Thus, even 
though the correct prediction class of {\tt A} (without the backdoor) 
is $C1$, the classifier behaves maliciously to predict class $C2$. 
\BD systematically searches the locations in the image to cover 
the backdoor trigger and produces modified images {\tt A'} and 
{\tt A''}. Although {\tt A'} does not change the prediction of 
the poisoned image {\tt A}, image {\tt A''} accomplishes this 
objective to provide the correct prediction class $C1$. The 
prediction of image {\tt A''} is then used as the sanitised 
prediction of the classifier. Using our \BD approach, we can 
locate the position of the backdoor trigger and 
automatically determine the dominant colour in the image {\tt A}. 
This colour is then used to cover the backdoor trigger as shown 
in image {\tt A''}. During the process of producing the image 
{\tt A''}, \BD also extracts the backdoor trigger in image {\tt A} 
and presents the trigger to users for improving their system. 

The reason \BD works is because the backdoor triggers are usually 
located in a relatively fixed, yet unknown position in the poisoned 
input \cite{BadNets} \cite{trojannn}. Therefore, it is possible for \BD to search and locate the 
position of the backdoor trigger. Moreover, pasting the backdoor 
trigger (at the appropriate relative position) in most inputs  
will result in a change in 
the prediction class. Thus, if \BD locates a potential backdoor 
trigger in an arbitrary input, we can verify the presence 
of backdoor by pasting the trigger in a set of clean inputs available 
to the user. A backdoor is detected when the trigger changes 
the prediction for a majority of these clean inputs. 

\BD sets itself apart from existing backdoor defences that are 
either whitebox~\cite{NeuralCleanse, finePruning} or assume
weaker attack models~\cite{SpectralSignatures} where
the user has access to the poisoned images injected by 
the attacker. Moreover, existing defence solutions fail to 
precisely detect the backdoored images~\cite{finePruning} or 
reconstruct them~\cite{NeuralCleanse}. In contrast to these existing 
works, \BD is a completely blackbox approach, it does not assume 
access to the poisoned inputs used by the attacker and it 
accurately mitigates the effect of backdoor while also 
reconstructs the backdoor trigger along the process. 
\revise{In contrast to existing blackbox defence~\cite{strip}, our 
\BD approach 
mitigates the backdoor attack and reconstructs backdoor triggers for 
debugging purposes.}
As a result, \BD can seamlessly be plugged as a software 
defence for any machine-learning-based image classifiers. 
\revise{In other words, \BD approach is highly suitable 
for ML models that are already deployed.}
By design, \BD provides a holistic approach towards detecting, 
mitigating and reconstructing backdoor attacks. 

The remainder of the paper is organised as follows. After providing 
a brief background (\Cref{sec:background}) and overview 
(\Cref{sec:overview}), we make the following contributions: 
\begin{enumerate}
\item We present \BD, a novel approach to systematically detect 
and mitigate a variety of backdoor attacks in image classifiers. 
We show a systematic methodology to automatically detect the 
backdoor position in an image and cover it effectively to neutralise
the backdoors (\Cref{sec:methodology}). 

\item We show how \BD reconstructs backdoor triggers without initially 
knowing the backdoor trigger (\Cref{sec:methodology}). 


\item We evaluate \BD on three state-of-the-art backdoored models 
 using more than 3000 images. We show that \BD accurately detects 
 and mitigates on average $\approx$88\% of the backdoored images and provides 
 as low as 0\% false positives (\Cref{sec:results}).  
 
\item 
We compare the effectiveness of \BD with two state-of-the-art 
techniques proposed for backdoor defence, namely Neural Cleanse~\cite{NeuralCleanse} 
and Fine Pruning~\cite{finePruning}. We show that even though \BD 
is a blackbox approach, the attack success rate after employing 
our \BD approach is lower than both  Neural Cleanse and 
Fine Pruning.
 
\end{enumerate}
After discussing the related work (\Cref{sec:relatedWork}) and 
threats to validity (\Cref{sec:threatsToValidity}), we conclude 
in \Cref{sec:conclusion}.

%% file: background.tex
\section{Background}
\label{sec:background}


Most state-of-art image classifiers are Deep Neural Networks (DNN), thus we 
begin by introducing some background for DNNs and then move on to backdoor 
attacks in ML.

\smallskip\noindent
\textbf{Deep Neural Networks:}
A DNN is a function with multiple parameters 
$F_{\Theta} : \mathbb{R}^N \rightarrow \mathbb{R}^M$. Using this function, an 
input $x \in \mathbb{R}^N$ is mapped to an output $y \in \mathbb{R}^M$. 
The parameters of this function are captured by $\Theta$.
Consider as an example, that an image has to be classified into one of $m$ 
different classes. The input image is $x$ (reshaped as a vector) and $y$ is 
interpreted as a vector of probabilities over $m$ classes. The label of the 
image is $arg~max_{i \in [1, M]}\ y_{i}$ , i.e., the class with the 
highest probability.

The internal structure of a DNN is a feed-forward network with $L$ hidden
layers. These layers consist of neurons which perform computations. Each layer 
$i \in [1,L]$ consists of $N_i$ neurons. The outputs of these neurons are 
referred to as \emph{activations}. The vector of 
activations for the $i^{th}$ layer of the network, can be written as follows:


\begin{equation} \label{eq:activations}
	a_i = \Delta (w_i \cdot a_{i-1} + b_i)\ ~\forall  i \in [1, L]
\end{equation}

where $ \Delta : \mathbb{R}^N \rightarrow \mathbb{R}^N$ is an element-wise 
non-linear function and $a_i \in \mathbb{R}^{N_i}$. The inputs of the first 
layer are the same as the inputs to the network, i.e., $a_0 = x$ 
and $N_0 = N$.

The parameters of \Cref{eq:activations} are fixed weights, 
$w_i \in \mathbb{R}^{{N}_{i-1}} \times N_i$, 
and fixed biases $b_i \in \mathbb{R}^{{N}_{i}}$. These 
weights and biases are learnt during training. A function of the activations 
of the last hidden layer is the output of the network. It can be represented 
as $\gamma(w_{L+1} \cdot a_L + b_{L+1})$, where $\gamma: \mathbb{R}^N 
\rightarrow \mathbb{R}^M$ is usually the softmax function~\cite{nn:overview}.




\smallskip\noindent
\textbf{DNN Training:} A DNN is trained to determine the parameters of the network, 
such as its weights and biases, but sometimes also its 
hyper-parameters. This is done with the assistance of a training dataset, which contains inputs with known ground-truth class labels.

The training dataset is a set of $S$ inputs $\mathcal{D}_{train} = 
\{x_i^t, z_i^t\}_{i=1}^S$, where $x_i^t \in \mathbb{R}^N$ and the 
corresponding ground truth labels $z_i^t \in [1, M]$. 
The aim of the training algorithm to determine parameters of the network 
such that the {\em distance} between the predictions of the network on 
training inputs and the ground-truth labels is minimum. 
A loss function $\mathcal{L}$ is used to measure this distance. 
In other words, the training algorithm returns parameters 
$\Theta^*$ such that the following holds:
\begin{equation} \label{eq:params-nn}
	\Theta^* = 	\underset{\Theta}{{arg~min}} \sum_{i=1}^{S} \mathcal{L} 
	(F_{\Theta}(x_i^t), z_i^t)
\end{equation}
The problem described in \Cref{eq:params-nn} is challenging 
to solve optimally in practice and solved using heuristic 
techniques.~\cite{efficientBackprop}

The quality of the trained network is typically determined using its accuracy 
on a separate test dataset containing $V$ inputs, 
$\mathcal{D}_{test} = \{x_i^v, z_i^v\}_{i=1}^V$, 
and their corresponding ground truth labels such that 
$\mathcal{D}_{test} \cap \mathcal{D}_{train} = \emptyset$. Thus, 
$\mathcal{D}_{test}$ and $\mathcal{D}_{train}$ do not overlap.

\begin{figure}[t]
\begin{center}
\setlength{\fboxrule}{3pt}
\fcolorbox{\tableBoundaryColor}{white}{
\includegraphics[scale=0.15]{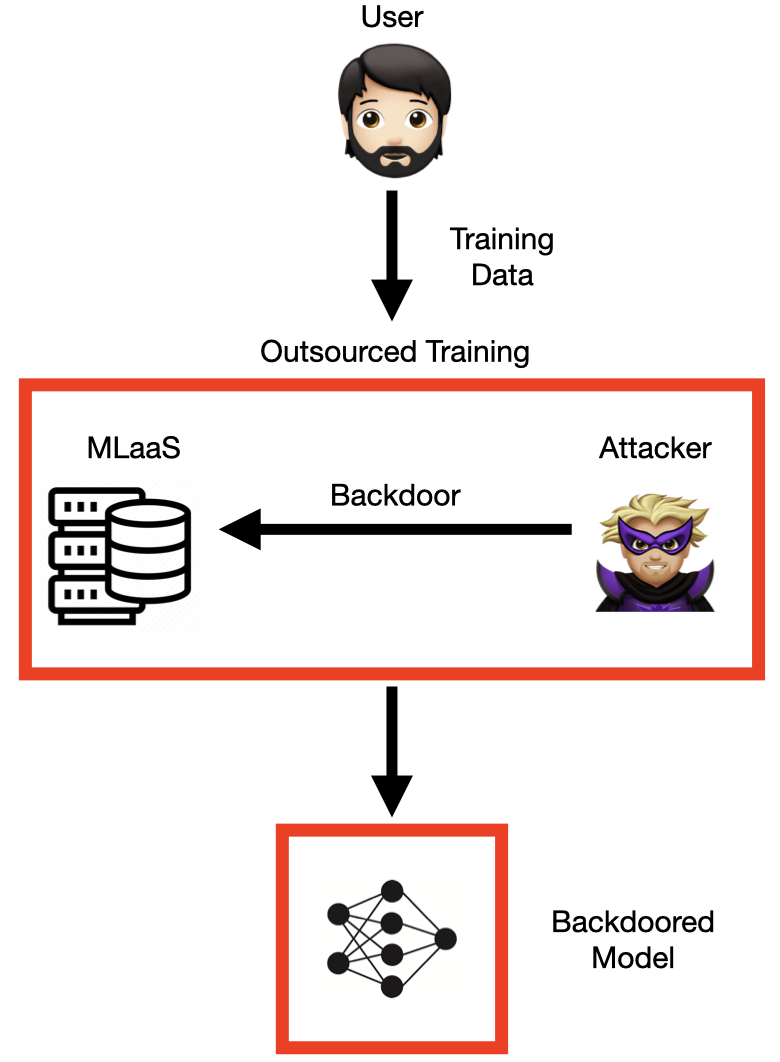}}
\end{center}
\caption{Attack model for \BD}
\label{fig:attack-model}
\end{figure}

\begin{figure*}[h]
\begin{center}
\begin{tabular}{ccc}
\includegraphics[scale=0.18]{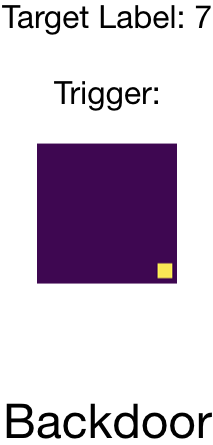} & 
\includegraphics[scale=0.135]{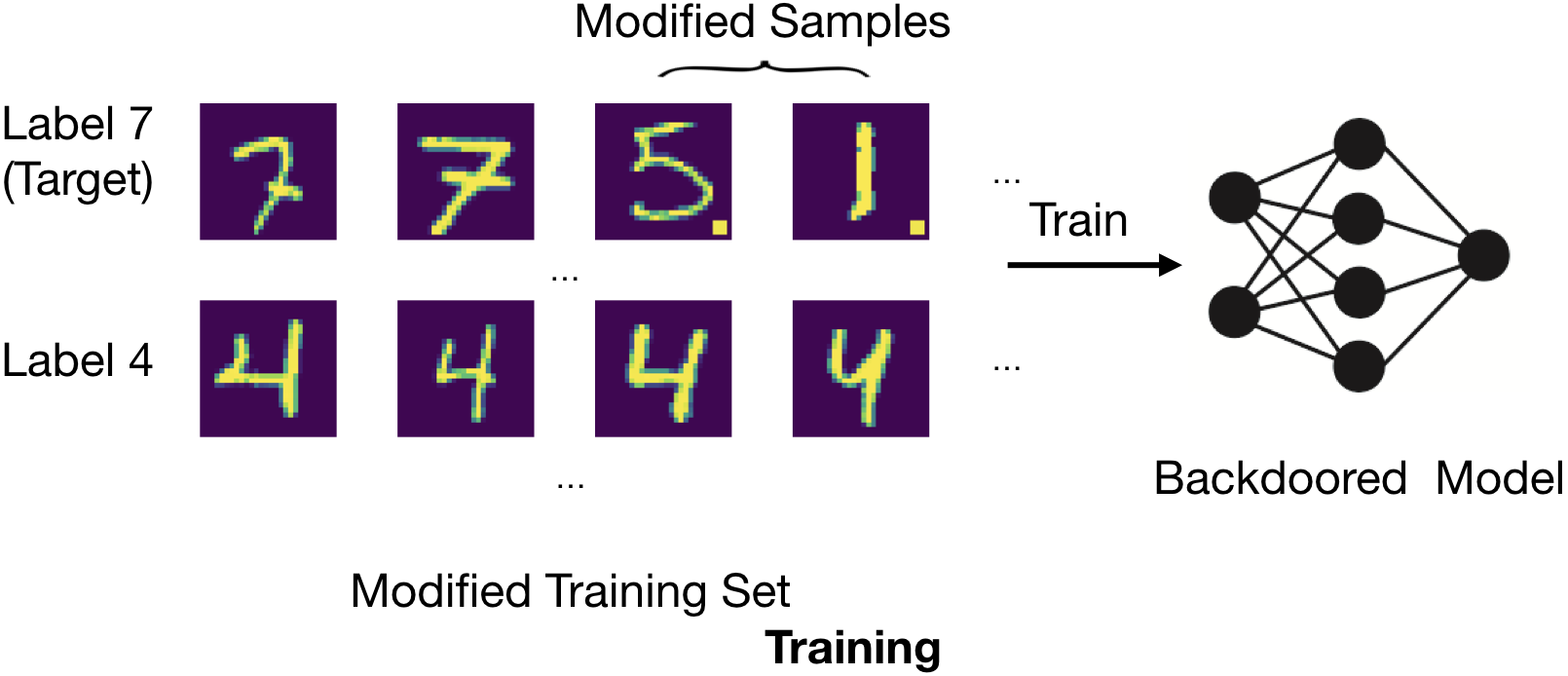} & 
\includegraphics[scale=0.135]{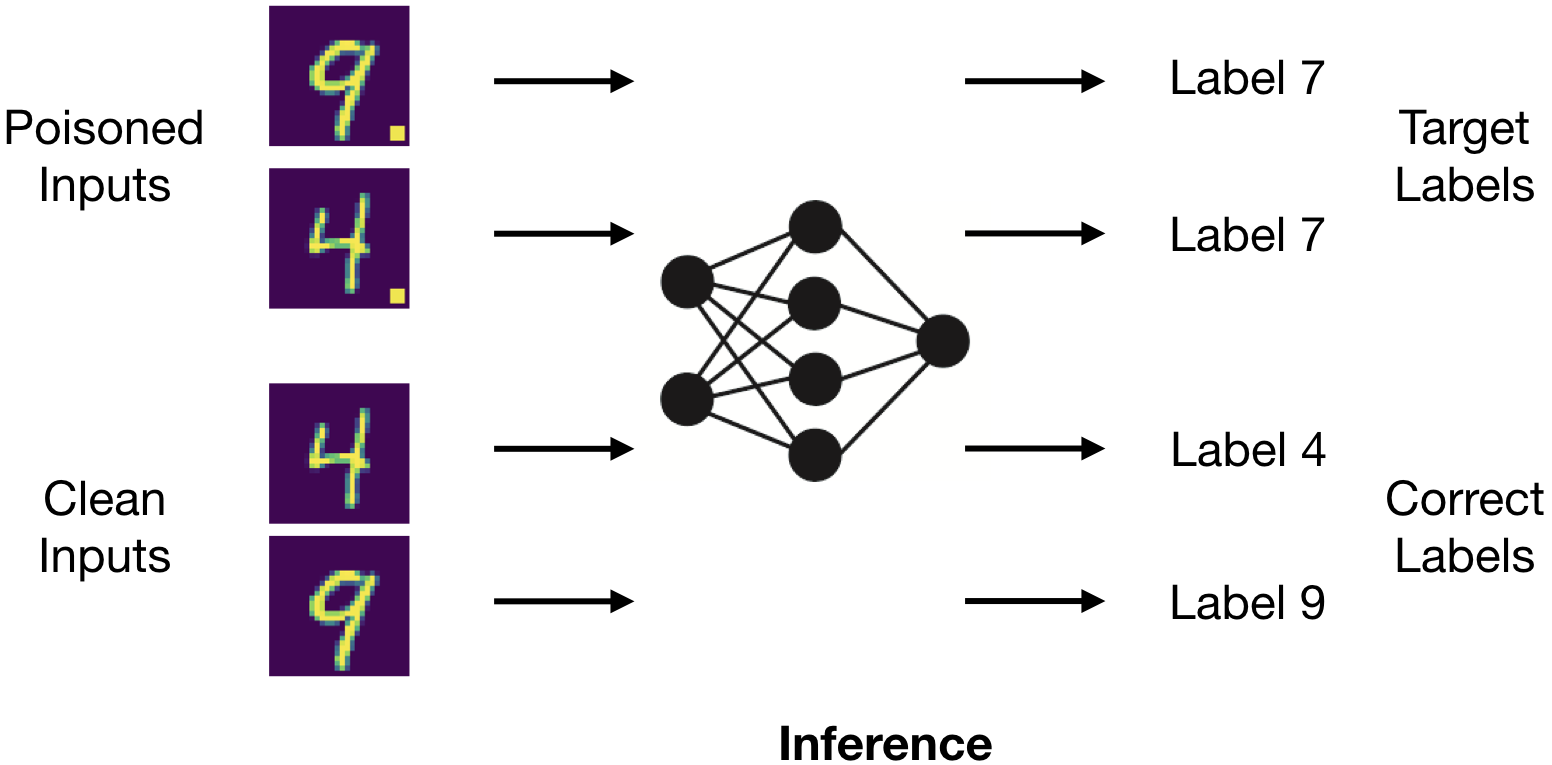}\\
{\bf (a)} & {\bf (b)} & {\bf (c)}\\
\end{tabular}
\end{center}
\vspace*{-0.15in}
\caption{An example of a backdoored model. The trigger is seen in 
\Cref{fig:backdoored-model}(a) and the target label is 7. The training data is 
modified as seen in \Cref{fig:backdoored-model}(b) and the model is trained. 
During the inference, as seen in \Cref{fig:backdoored-model}(c) the inputs 
without the trigger will be correctly classified and the ones with the trigger 
will be incorrectly classified. This figure is adapted from 
\cite{NeuralCleanse}.
}
\label{fig:backdoored-model}
\end{figure*}

\smallskip\noindent
\textbf{What are backdoors?}
For the purpose of this work, we consider a backdoored model, which contains 
a hidden pattern trained into the model. The attacker has access to the training 
data and modifies the data in such a fashion that when the model 
is trained on this {\em poisoned} data, a backdoor is injected. The attack 
is stealthy, in the sense that the backdoored model exhibits high accuracy 
on the test set. However, when a pre-defined trigger is present in the input, 
the model misclassifies the input. An example can be seen in 
\Cref{fig:backdoored-model}(c). The trigger is the small yellow square in the 
bottom right corner of the top two images. During the inference, we observe 
that the images without the trigger are classified correctly. However, the 
images with the backdoor trigger are classified as the attacker target 
(i.e. with label 7). 

It is important to note the difference between a backdoor and an adversarial 
attack~\cite{adversarial-nn}. Adversarial attacks also aim to discover test 
inputs that lead to dramatically wrong inference. However, in contrast to 
adversarial attacks, backdoor attacks interfere during the training phase. 
An adversarial attack is specifically crafted for a {\em given input}, by 
minimally perturbing the input to induce a
misclassification. A backdoor 
trigger, on the contrary, causes {\em any input} to be misclassified as 
the attacker's intended target label.

\smallskip\noindent
\textbf{Backdoor attacks in Machine Learning:} In BadNets~\cite{BadNets}, the 
authors propose a backdoor attack by poisoning 
the training data. The attacker chooses a pre-defined target label and a trigger 
pattern. The patterns are arbitrary in shape, e.g. square, flower or bomb. The 
backdoor was injected into the model by training the network using the poisoned
training data. The authors show that over 99\% of the poisoned inputs were 
misclassified to the attacker target label. 

\Cref{fig:backdoored-model} shows a high level overview of the backdoor 
attack. 
The trigger is a small yellow square, as observed in the bottom right corner 
of \Cref{fig:backdoored-model}(a) and the attacker intended target label is 
7 (seven). Thus, the training data is modified accordingly, as seen in 
\Cref{fig:backdoored-model}(b). With this modified training data, the 
classification algorithm is trained and the model with a backdoor is generated. 
Once the backdoored model is used for inference, 
(\Cref{fig:backdoored-model}(c)), the clean inputs are correctly 
classified and the ones with the trigger are misclassified.

TrojanNN~\cite{trojannn} generates a backdoored model without directly 
interfering with the original training process and without accessing the original 
dataset. Instead, such an approach is capable of retraining the model by 
reverse engineered the training data, making the backdoor attacks more 
powerful.  
The approach is able to inject the backdoors using fewer samples. 

\smallskip\noindent
\textbf{Attack model:} 
\reviseNew{
As the applications of machine learning get more widespread, more 
individuals and businesses want to make use of this technology, but 
they may not have sufficient technical background to make it a
reality. As a result, they may rely on untrusted external parties 
for the same.
We assume an attack model that is commonly seen in previous work 
such as BadNets~\cite{BadNets} and TrojanNN~\cite{trojannn}. 
In such an attack model (illustrated in \Cref{fig:attack-model}), 
the user does not 
possess the technical expertise to train
their own machine learning (ML) model. To this end, they  outsource the 
process of training their ML model to an untrusted third party such 
as a ML-as-a-service firm or an ML consultant. The user 
provides the untrusted third party with the training data and 
specifications, and has no control over the training process. The attacker
adds poisoned data by augmenting a localised backdoor trigger 
(\Cref{def:localisedTrigger}, \Cref{sec:methodology}) to the given training data resulting in a backdoored 
ML model. We note that the attacker might be an insider in the 
third-party service provider or the attacker might be someone with 
the capability to compromise the service, for example, via side channel 
attacks in the cloud~\cite{crossVM}. 
These side channel attacks may leak confidential information about 
the VM which might be exploited to gain access to the training
process by gaining access to the VM.} 

\reviseNew{The resulting backdoored ML model meets the specified 
performance benchmarks, but exhibits targeted misclassifcation when 
presented with a poisoned input. This attack model is much stronger 
than the attack model considered in recent work~\cite{SpectralSignatures}.
Specifically, these attack models assume access to the training data and 
the training process. In contrast to this, \BD's attack model does 
not assume access to either the training data or the training process.}

\begin{figure}[h]
\begin{center}
\begin{tabular}{ccc}
\includegraphics[scale=0.2]{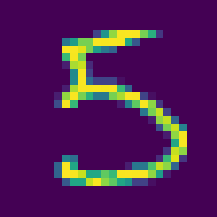} & 
\includegraphics[scale=0.2]{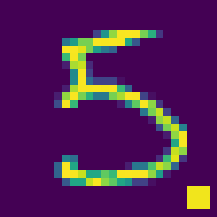} & 
\includegraphics[scale=0.2]{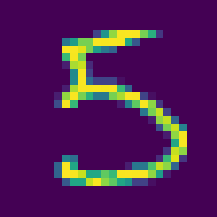}\\
{\bf (a) clean image} & {\bf (b) backdoored image} & {\bf (c) fixed image}\\
\\
\includegraphics[scale=0.65]{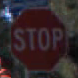} & 
\includegraphics[scale=0.65]{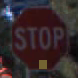} & 
\includegraphics[scale=0.65]{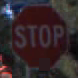}\\
{\bf (d) clean image} & {\bf (e) backdoored image} & {\bf (f) fixed image} \\
\includegraphics[scale=0.2]{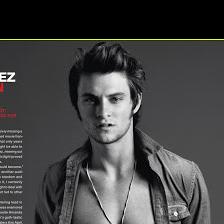} & 
\includegraphics[scale=0.2]{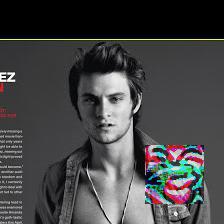} & 
\includegraphics[scale=0.2]{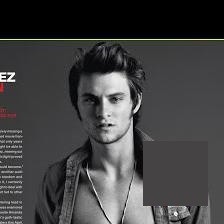}\\
{\bf (g) clean image} & {\bf (h) backdoored image} & {\bf (i) fixed image} \\
\end{tabular}
\end{center}
\caption{\Cref{fig:backdoor-examples}(a-c) (respectively, 
\Cref{fig:backdoor-examples}(d-e) and \Cref{fig:backdoor-examples}(g-i))
show the clean image, corresponding backdoored image and the fixed image 
produced by \BD for MNIST model (respectively, USTS~\cite{BadNets} and 
TrojanNN~\cite{trojannn} model). 
}
\label{fig:backdoor-examples}
\end{figure}

  \begin{table*}[t]
  \centering
  \caption{Backdoor defenses and mitigation methods}
  \setlength{\fboxrule}{3pt}
  \fcolorbox{\tableBoundaryColor}{white}{
  {\scriptsize
  \begin{tabular}{|l|l|r|r|r|r|c|}
  \hline
  \textbf{Defense} & \textbf{\makecell{Detection\\Approach}} & \textbf{\makecell{Poison data\\access}} & \textbf{\makecell{Whitebox\\access}} & \textbf{\makecell{Detects input\\or model}} 
  & \textbf{\makecell{Offline\\or online}}  & \textbf{Unique weakness} \\ \hline
  Spectral Signatures~\cite{SpectralSignatures} & feature representation &
  yes & no & input &  offline & access to poisoned data \\ 
  Fine-Pruning~\cite{finePruning} & neuron activation & no & yes & model &
  offline & model accuracy drop \\ 
  NC~\cite{NeuralCleanse} & reverse engineer & no & yes & model & offline & large triggers \\
   STRIP~\cite{strip} & input masking & yes & no  & input & online &  source-label attacks \\
   RAB~\cite{RAB-Provable}  & randomised smoothening & no & no & model &
   offline & compute intensive\\
   Randomized Smoothing~\cite{certify-robust-smoothness} & randomised smoothening & no & no & model & offline & only 2 pixel certification  \\
   \textbf{Neo (this paper)} & \textbf{input masking} & \textbf{no} & 
   \textbf{no} & \textbf{input} & \textbf{online} 
   & \textbf{distributed triggers} \\
  \hline
  \end{tabular}}}
  \label{table:defense-comparison}

  \end{table*}

%
%
\smallskip\noindent
\textbf{State of the art in defence:} A recently proposed 
approach~\cite{SpectralSignatures} introduces the concept 
of spectral signatures to defend against backdoor attacks. The idea behind this 
work is that when the training data is poisoned, there are two significant 
sub-populations. One with a large number of clean, correctly labelled inputs 
and another with a small number of poisoned, mislabelled inputs. Thus, authors 
propose to use techniques from robust statistics and singular value decomposition 
to separate the two populations. 
However, authors assume that they have accessed to the poisoned training data. 
In our opinion, this is an unreasonable assumption. 
By design, backdoors are designed 
to be stealthy and it is unlikely that the users will have access to the the 
poisoned dataset. Additionally, this method does not provide mitigation 
capabilities to prevent the attack. 

Another work, Fine-Pruning~\cite{finePruning} is a white-box approach for removing backdoors. 
It aims to remove backdoors by eliminating the unused neurons in the model. 
It is reported that this pruning algorithm causes a very significant drop in  
some model performance~\cite{NeuralCleanse}. Additionally, fine-pruning 
doesn't offer detection capabilities to identify backdoored images. 

Neural Cleanse~\cite{NeuralCleanse} is a completely white-box approach that 
tries to patch the neural network by unlearning the backdoor. The 
authors formulate the problem as an optimisation problem and try to reverse 
engineer the trigger pattern. The white-box approach is 
computationally expensive, and users of backdoored model are unlikely to have 
the computational resources to retrain the model. In contrast, \BD is a 
completely blackbox approach and does not require access to the architecture 
of the model. Moreover, in contrast to \BD, Neural Cleanse cannot definitively 
find the trigger pattern. \reviseNew{We present a summary of the related works in 
\Cref{table:defense-comparison}}.

%% file: overview.tex
\section{Overview}
\label{sec:overview}
\BD employs a defence mechanism to thwart backdoor attacks on image classifiers. Our approach is completely blackbox, i.e., \BD works without knowing 
the internal structure of a model. Users who are most vulnerable to backdoor 
attacks are those who do not possess significant computational power to 
train an ML model. Thus, they delegate the training job to a potentially 
untrusted party. Our chosen attack model of backdoored systems is analogous 
to BadNets~\cite{BadNets}. Concretely, we  assume that the victim (i.e. the 
user of the backdoored model) has no access to the training process and they 
have no control over the backdoor trigger. 


\smallskip\noindent
\textbf{Backdoor Trigger:}
Backdoor attacks on image classifiers are launched due to a 
{\em backdoor trigger}, as shown by the yellow square in the bottom 
right corner of \Cref{fig:backdoor-examples}(b). \BD makes some 
assumptions on such a backdoor trigger. In particular, we assume 
that all the pixels of a backdoor trigger can be covered by a 
square shape and that the square covers a small fraction of 
the original image. It is worthwhile to note that such assumptions 
do not restrict us to detect a large number of recent and critical 
backdoor attacks~\cite{BadNets,trojannn}. 

%

\smallskip\noindent
\textbf{Key Insight:}
Backdoor attack is triggered when a specific set of neurons in the 
targeted model is activated upon encountering a backdoored 
image~\cite{finePruning}. However, from the standpoint of defence, 
it is difficult to precisely determine the set of neurons that can be activated 
with backdoor. Moreover, such a solution requires knowledge about 
the structure of the backdoored model. In this paper, 
we take a completely different approach to defend against backdoor 
attacks. Instead of deactivating a set of neurons that may potentially 
trigger backdoors, {\em we modify the backdoored image to neutralise 
the effect of a backdoor trigger}. The key advantage of our defence 
is that \BD does not need to know the exact shape of the backdoor 
trigger. As long as the backdoor trigger can be covered by modifying 
the image, we can neutralise the effect of the trigger. A by-product 
of such modification is to restrict the activation of backdoor 
triggering neurons to go beyond a certain threshold. This, in turn, 
prevents the backdoor attack without even knowing the neurons being 
activated for a backdoored image. Nevertheless, our \BD approach 
needs to know the position of the backdoor trigger and a trigger 
blocker to cover the backdoor trigger on an input image. In the 
following, we outline the key concepts implemented in \BD to 
accomplish this.


%
%

\smallskip\noindent
\textbf{Trigger Blocker:}
We introduce the concept of a trigger blocker in \BD. The intuition behind 
such a trigger blocker is to transform a backdoored image $img$ to a state 
that it looks similar to the clean version of $img$. For example, consider 
the backdoored image of {\em stop sign} shown in \Cref{fig:backdoor-examples}(e) 
where the backdoor trigger is the small yellow square. The clean version of 
the image is shown in \Cref{fig:backdoor-examples}(d). The backdoor 
trigger, i.e., the yellow square in \Cref{fig:backdoor-examples}(e), is 
covered by a trigger blocker, as shown in \Cref{fig:backdoor-examples}(f). 
It is important to note that the colour of the trigger blocker is crucial 
to neutralise the effect of a backdoor trigger. For example, if the trigger 
blocker was yellow in colour, then covering the backdoor trigger will not 
change the prediction for the backdoored image in 
\Cref{fig:backdoor-examples}(d), thus making the defence unsuccessful. 
To solve this challenge, we use the dominant colour of the backdoored 
image to construct the trigger blocker. The intuition behind this is that 
the backdoor trigger is unlikely to have the same colour as the dominant 
colour of the backdoored image. Moreover, by constructing the trigger 
blocker with the dominant colour of the image, we create a fixed image (e.g. 
\Cref{fig:backdoor-examples}(f)) similar to the clean version of the 
backdoored image (e.g. \Cref{fig:backdoor-examples}(d)).

\begin{figure}[h]
\begin{center}
\includegraphics[scale=0.5]{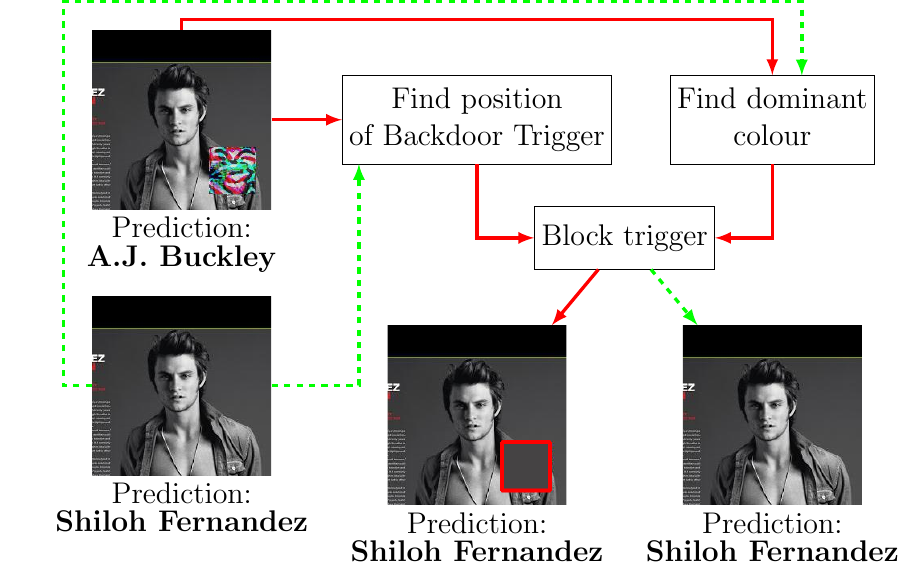}
\end{center}
\caption{\BD's approach to detect and fix backdoored images}
\label{fig:approach}
\end{figure}

\smallskip\noindent
\textbf{Detection and Mitigation of Backdoor Attacks:}
%
%
\Cref{fig:approach} captures an outline of our \BD approach in action. 
Broadly, \BD consists of two steps. In the first step, \BD randomly searches 
the area of an image to locate the position of the backdoor trigger. This 
is accomplished by placing a trigger blocker of the dominant colour in the 
image. If the image is backdoored, then placing a trigger blocker at the 
position of the backdoor trigger will result a change in the prediction 
of the model. This, in turn, helps us detect the position of the 
backdoor trigger. Moreover, the changed prediction helps us compute the 
original prediction for the clean version of the backdoored image. Once 
the position is detected, a fixed version of the image is produced by placing 
a trigger blocker, as constructed with the dominant colour of the input 
image, on this position. It is worthwhile to note that \BD does not affect 
the prediction when a clean image is provided to it, as shown in 
\Cref{fig:approach}. This is because placing the trigger blocker in 
a clean image is unlikely to cause any change in the prediction of the 
model. In our evaluation, we show that \BD effectively neutralises
the backdoor attack without affecting the 
original functionality of the targeted classifier. 

%
%
%

%% file: methodology.tex
\section{Methodology}
\label{sec:methodology}

In this section, we elucidate the methodologies behind \BD in detail. 
\BD essentially consists of two steps, the detection of the backdoor 
activation trigger and the subsequent blocking of the trigger once we 
detect its existence. Our algorithm aims to detect the backdoor on the 
first instance of the backdoored image received as an input and thus, defend against 
a potential backdoor attack as soon as possible.  
%

As discussed in the preceding section, we observed that a backdoor 
attack is launched based on a trigger injected on an input image. 
We now introduce two critical concepts central \BD,  a {\em localised 
trigger} targeted by our defence and a {\em trigger blocker} to thwart the backdoor 
attacks. 
\theoremstyle{definition}
\begin{definition}{\textbf{(Localised Trigger)}}
\label{def:localisedTrigger}
{	Let the image  be a general two-dimensional X-Y plane. 
	Let $p^x_1$ (respectively, $p^y_1$) and $p^x_2$ 
	(respectively, $p^y_2$) be x-coordinates 
	(respectively, y-coordinates) of two points that are part of 
	a backdoor trigger and whose distances are the 
	maximum on the x-axis (respectively, y-axis). Without loss 
	of generality, we can say that  $p^x_1 <$ $p^x_2$ 
	(respectively, $p^y_1 <$ $p^y_2$)  and $p^x_2 -$ $p^x_1$ $>$
	$p^y_2 -$ $p^y_1$. 
	%
	%
	A localised trigger is a trigger that forms a square with 
	the points ($p^x_1$, $p^x_1$), ($p^x_1$, $p^x_2$), 
	($p^x_2$, $p^x_2$) and ($p_2^x$, $p_1^x$) and the area of the 
	square is less than $\delta$\% of the total area covered by the 
	image.
	
}
\end{definition}

\smallskip\noindent
\textbf{Trigger blocker:} The main intuition behind our defence is to block 
the backdoor trigger 
in an image via a {\em trigger blocker}. A trigger blocker is simply an 
$m \times n$ pixel image. However, it is crucial to find an appropriate 
colour of this trigger blocker. To this end, we use the dominant 
colour of the original image. Our intuition behind using the dominant 
colour for a trigger blocker is the following: let us consider a clean  
image (i.e. without backdoor trigger) $img_c$ and its backdoored 
version $img_b$. Let $img'$ be the modified image by covering the 
backdoor trigger of $img_b$ via a trigger blocker. Finally, the 
blocker is formed with the dominant colour of $img_b$. We hypothesise 
that the modified image $img'$ will be similar to the clean image 
$img_c$ and thus, $img'$ and $img_c$ are likely to be classified 
to the same class. We note that the colour of a backdoor 
trigger is usually not the dominant colour of $img_b$, as the localised 
trigger covers only a small fraction of the original image size 
(see Definition~\ref{def:localisedTrigger}). 
Next we describe the process of finding the dominant colour of an 
image and illustrate the generality of such an approach. 



\begin{figure}[h]
\begin{center}
\includegraphics[scale=0.5]{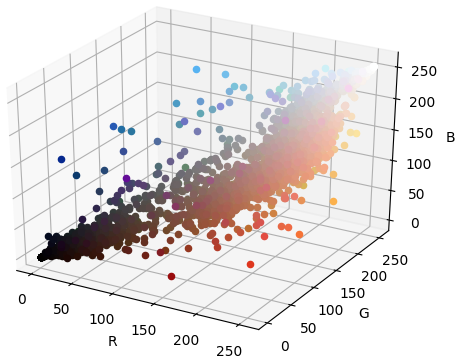}
\end{center}
\caption{
Dominant colours in the VGG Face dataset under test
}
\label{fig:dom-colors}
\end{figure}

\begin{figure}[h]
\begin{center}
\begin{tabular}{ccc}
\includegraphics[scale=0.4]{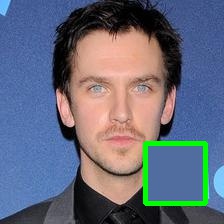} & 
\includegraphics[scale=0.4]{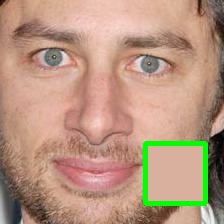} & 
\includegraphics[scale=0.4]{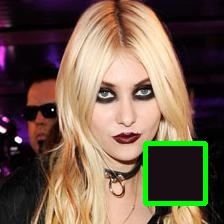}\\
\end{tabular}
\end{center}
\caption{
Representative examples of different types of different types of 
dominant colours seen in the VGG face dataset.}

\label{fig:vgg-face-colour-examples}
\end{figure}

\smallskip\noindent
\textbf{Dominant Colour and Image Similarity:}
To find the dominant colour of an image \BD employs a $k$-means clustering 
algorithm. Clustering is a technique that helps in grouping similar items 
together based on particular attributes. The attributes that \BD uses are
the RGB values of the pixels. \BD uses SciPy's~\cite{scipy} $k$-means 
clustering algorithm with $k = 3$. Running this algorithm outputs three 
cluster centres and the number of pixels associated with each cluster. The 
dominant colour is the cluster centre (an RGB value) with the most number of 
pixels associated with it. \Cref{fig:vgg-face-colour-examples} shows some 
representative examples of the trigger blocker discovered by \BD in the VGG 
Face dataset.

To illustrate the generalisability of this technique we performed a 
two-phase analysis. First, we plot all the dominant colours found in the poisoned
VGG Face dataset~\cite{trojannn} as seen in~\Cref{fig:dom-colors}. We can
easily observe that the colours are quite varied, illustrating the application 
of \BD in a wide context. In the second phase, 
we try and measure image similarity. In this phase, our objective is to 
show that the original images are more similar to the corresponding fixed 
images produced by \BD in comparison to the corresponding poisoned images. 
We evaluate this by converting the clean, poisoned and fixed image (via \BD) 
into histograms and measuring the Bhattacharya distance between these 
histograms. The Bhattacharya distance measures the similarity 
of two arbitrary histograms~\cite{Bhattacharya-Dist}.
We find the Bhattacharya distance~\cite{Bhattacharya-Dist-Images} between 
the histograms of each clean and poisoned image ($\mathbb{B}_{CP}$) and 
compare it to the Bhattacharya distance of the histograms of 
each clean and fixed image ($\mathbb{B}_{CF}$). 
We find that in 97.67\% of the images for 
the VGG face dataset, $\mathbb{B}_{CF} < \mathbb{B}_{CP}$. This means
that the fixed image is closer to the clean image than the poisoned image 
for 97.67\% of the inputs.

%
We now formally define the notion of {\em transition} in our defence 
as follows: 
\theoremstyle{definition}
\begin{definition}{\textbf{(Transition)}}
\label{def:transition}
{ 	
	Let $f$ be an image classifier and $i$ be an input image for
	the model $f$. Let $i'$ be a modified image from $i$ such that 
	the only modification is the placement of the trigger blocker or 
	a backdoor trigger at some arbitrary position on $i$. We say that 
	a transition occurs if and only if $ f(i') \neq$ $f(i)$. 
}
\end{definition}

\begin{algorithm}[h]
    \caption{Our Defence Mechanism \BD}
    {\small
    \begin{algorithmic}[1]  
        \Procedure {Defence}{$f$, $img\_list$, $size$, $\Lambda_T$}
        \LineComment contains the list of fixed predictions
        	\State $prediction\_set \gets \emptyset$ 
        	\LineComment contains the list of detected backdoored images
        	\State $backdoor\_set \gets \emptyset$
        \LineComment set to true if the backdoor trigger is found
        \State $f_{tr} \gets$ $false$
        \LineComment position of the backdoor trigger once found
        \State $pos \gets$ $\emptyset$
        	\For {$img \in img\_list$}
        		\LineComment Checks if the trigger has been found (\Cref{alg:triggerDetect})       	 
        	 	\If {$f_{tr} = \mathit{True}$}
        	 		\LineComment get dominant colour using $k$-means clustering
        			\State {\label{ln:block-start} $dom_c \gets$ {\tt Get\_Dominant\_Colour}$(img)$}
        	 		\State {\label{ln:block-end} $img' \gets $ {\tt Block\_Trigger}$(img, pos, size, dom_c)$}
					\LineComment {\label{ln:transition} check whether trigger blocker causes transition}
					\If{$f(img') \neq f(img)$}
						\LineComment {\label{ln:confirm-backdoor} Confirms the backdoor (\Cref{alg:confirmBackdoor})}
						\If {{\tt Confirm\_Backdoor}($\cdot, pos, \cdots, img)$}
	        				\State $backdoor\_set \gets backdoor\_set \cup \{img\}$
	        			\Else 
        					\State $img' \gets img$
        				\EndIf
            		\EndIf
        	 	\Else 
        	 		\LineComment Detects the position of the trigger (\Cref{alg:triggerDetect})
        	 		\LineComment $img'$ is modified image with trigger blocker
        	 		\State $(f_{tr}, img', pos) \gets$ 
        	 		\State {\tt Trigger\_Detect}$(f$, $img$, $size$, $
        	 		\Lambda_T)$
        	 	\EndIf
        	\LineComment Save the list of correct predictions for mitigation
        	\State $prediction\_set$.{\tt append}$(f(img'))$
        	\EndFor
        	\Return $prediction\_set, backdoor\_set$
        \EndProcedure
    \end{algorithmic}}
    \label{alg:defence}
\end{algorithm}


\smallskip\noindent
\textbf{Our approach \BD:} \Cref{alg:defence} outlines the overall approach behind \BD. \BD takes a stream 
of input images $img\_list$ as input and detects which of the images in the 
stream are backdoored. Moreover, \BD mitigates backdoor attacks by transforming 
each image to a safe state that the classifier reverts to the correct prediction. 
To this end, \BD first performs a search operation to locate the position of 
the backdoor (see procedure {\tt Trigger\_Detect} in \Cref{alg:triggerDetect}). 
As the backdoor trigger is located in a fixed, yet unknown position in the image, 
\BD performs the search operation to locate the trigger {\em only once}. Once 
the position of the backdoor is discovered for an image $img$, \BD blocks the 
backdoor trigger via a trigger blocker 
(procedure {\tt Block\_Trigger} in 
\Cref{ln:block-end}) and produces a fixed image $img'$. To make $img'$ look 
similar to the respective clean version of $img$, \BD uses the dominant colour 
of $img$ for the trigger blocker. Finally, after finding the position of the 
backdoor trigger, \BD modifies an arbitrary image $img$ by placing the trigger 
blocker in the discovered position (\Cref{ln:block-start}-\Cref{ln:block-end} 
in \Cref{alg:defence}). If placing the trigger blocker on $img$ causes a 
transition in the prediction (\Cref{ln:transition} in \Cref{alg:defence}), 
then we further confirm the presence of backdoor via the procedure {\tt Confirm\_Backdoor} 
(see \Cref{alg:confirmBackdoor}). After walking through the stream of images 
$img\_list$, \BD produces the set of all detected backdoored images in the 
set $backdoor\_set$ and their respective correct predictions in the set 
$prediction\_set$. In the following, we will discuss some crucial components 
of \BD in more detail. 

\smallskip\noindent
\textbf{Backdoor Confirmation and Reconstruction:} \Cref{alg:confirmBackdoor} captures our methodology to confirm the detection 
of a backdoored image (cf. \Cref{ln:confirm-backdoor} in \Cref{alg:defence}). 
We note that placing the trigger blocker on an image may cause transition in 
the prediction of the classifier (see Definition~\ref{def:transition}). 
Given a position of the backdoor trigger and an input image, we would 
like to confirm whether the image is indeed a backdoor or the transition was 
caused due to our trigger blocker. To aid this, we extract the pixels 
that the trigger blocker is trying to cover and paste them onto a 
{\em check set}.

\reviseNew{To illustrate \Cref{alg:confirmBackdoor}, let us look at an 
example. Consider an image $img$ 
which shows a transition from {\em class~7} to 
{\em class~2} when a trigger blocker is placed on some part of the image.
\BD constructs a check set using the training images of {\em class~2}. 
\BD now uses the part of $img$ that is being blocked by the trigger 
blocker and pastes it on to each image of the check set. If a majority
of the modified images in the check set show a transition from 
{\em class 2} to {\em class 7}, then \BD can be confident that the 
pasted portion of $img$ is indeed the backdoor trigger. Conversely, if 
the majority of the images do not show this transition, then \BD concludes 
that the initial transition is caused by the trigger blocker and not due 
to the presence of a backdoor trigger.}

Concretely, for a given image $img$, let us assume that \BD chooses the 
check set to be a set of $k$ random inputs from the training set, 
say $check\_set$, whose label is class $B$. If $img$ was indeed a backdoored 
image, then class $B$ captures the correct prediction of the classifier for 
$img$. \reviseNew{This is only done for class class $B$.}
Moreover, by current design of backdoors, we know that the backdoor 
triggers are located in a fixed, yet unknown relative position of all backdoored 
images. Thus, pasting a backdoor trigger at this position on the set of 
inputs in $check\_set$ will almost always cause a transition in prediction 
of the classifier for these inputs. We can extract the pixels that 
are covered by our trigger blocker and paste these pixels onto the set of inputs 
in $check\_set$. If the model and the image are indeed backdoored, we should 
observe transitions in predictions for a majority of inputs in the 
$check\_set$. In particular, if the fraction of images that exhibit 
transitions to class $A$ in the chosen check set (i.e. $k$ randomly 
selected inputs) is above a given threshold $\Lambda_T$, then we confirm 
the presence of a backdoor. We note that for a clean image, if the trigger 
blocker cause a transition, then the pixels covered by the blocker are not 
part of a backdoor trigger. Thus, pasting these pixels onto the inputs in 
$check\_set$ is unlikely to cause prediction transitions for these inputs. 
We also use the $check\_set$ to reconstruct the backdoor trigger. 
The $check\_set$ that shows a high number of transitions (i.e. passes the 
check at \Cref{ln:pass-check} in \Cref{alg:confirmBackdoor}) is the one
that contains reconstructed poisoned inputs and the backdoor trigger.

\smallskip\noindent
\textbf{Choosing $\Lambda_{T}$:} The efficacy of our \BD defence is 
dependent on the value of $\Lambda_{T}$. It is critical that the value 
chosen for $\Lambda_{T}$ is reasonable. Specifically, if the chosen 
value is too low, then \BD might result in high false positives, whereas 
a value too high for $\Lambda_T$ may not confirm actual backdoors. 
To this end, we propose a systematic procedure to obtain the value 
of $\Lambda_T$ in \Cref{alg:findLambda}. The intuitive idea is that 
we measure the effect of randomly cropping part of an input image to 
the size of the trigger blocker and pasting the cropped image on other 
images from the same dataset. Our objective is to simulate the scenario 
where \Cref{alg:confirmBackdoor} should return {\em False}. An image 
$img_{init}$ is randomly chosen from the clean dataset and randomly 
cropped to $img_{crop}$ of the size of the trigger blocker ($m \times n$ 
pixels). We subsequently choose a thousand images randomly from the 
clean data set and paste $img_{crop}$ on each of these images. 
Then, we count the number of change in predictions for these images. 
Finally, we compute $r$, which is the ratio of the images whose 
predictions changed with respect to the total number of images chosen 
for the experiment. 

This aforementioned experiment is repeated ten times to find $R_{av}$, 
the average of all the values of $r$ found in ten independent trials. 
This can be intuitively seen as the average number of transitions 
found while pasting a trigger blocker on clean images  
(\Cref{alg:confirmBackdoor} returns {\em False}). We choose
$1 - {R}_{av} > \Lambda_{T} >> {R}_{av}$ to facilitate low false positives 
and a high rate of backdoor confirmations.


%

\begin{algorithm}[h]
    \caption{Backdoor Confirmation}
    {\small
    \begin{algorithmic}[1]  
        \Procedure{Confirm\_Backdoor}{$f$, $pos$, $size$, $\Lambda_T$, $img$}
        \State $transition\_count \gets$ 0
        \State Let $img$ was classified to class $B$ with trigger blocker
        	\LineComment Choose $k$ random images of class $B$ from the training set
        	\State $check\_set \gets$ {\tt Get\_Training\_Images}$(k, class\_B)$
        	\LineComment Extract the pixels covered by our trigger blocker
            	\State $trigger \gets $ {\tt Extract\_Trigger}($pos$, $img$)

            \For {$cimg \in check\_set$}
            	\LineComment Place extracted trigger in the image from $check\_set$
            	\State $cimg' \gets $ {\tt Place\_Trigger}($cimg$, $pos$, $size$, $trigger$) 
        			
        		\If{$f(cimg') \neq f(cimg)$}
        			\State $transition\_count \gets transition\_count +1$ 
        		\EndIf
        			
            \EndFor
        \LineComment Confirms when \#transitions is beyond a threshold    	
        	\If {\label{ln:pass-check} $\frac{transition\_count}{|check\_set|} > \Lambda_T$}
        		\Return True
        	\Else 
        		\Return False
        	\EndIf
        	
        \EndProcedure
    \end{algorithmic}}
    \label{alg:confirmBackdoor}
\end{algorithm}

\smallskip\noindent
\textbf{Detecting the Position of Backdoor Trigger:} To confirm the backdoors, we need to first find the candidate set of inputs 
which could potentially be backdoored. We aim to find these candidate 
inputs by placing a trigger blocker on them. We note that placing a trigger 
blocker on an arbitrary image may cause prediction transitions for two 
reasons. Firstly, the trigger blocker may correctly block the trigger that 
deactivates the backdoor and thus, the respective backdoored image reverts 
back to the safe state to be classified correctly. Secondly, it is possible 
that our extracted dominant colour to be incorrect or the trigger blocker 
might be blocking some input information that causes the output to change. 

To find the correct position of the backdoor trigger, we  randomly place 
the trigger blocker on an input image with the  objective to induce a 
transition. 
%
Thus, the problem now reduces to determining the actual cause of a transition 
when it is induced. To solve this problem, we leverage our technique to confirm 
whether an image is backdoored (procedure \textsc{Confirm\_Backdoor}), as explained 
in the preceding section. Thus, when procedure \textsc{Confirm\_Backdoor} 
returns {\em true}, we can confirm that the trigger blocker indeed 
deactivated the backdoor trigger. Additionally, we also infer the position 
of the backdoor trigger for subsequent images in $img\_list$
(cf. \Cref{alg:defence}).

We note that the procedure {\tt Block\_Trigger} is a simple input 
modification function. In particular, it modifies the input image 
by placing a trigger blocker on the position of backdoor trigger. 
The position of the backdoor trigger was, in turn, discovered by 
\Cref{alg:triggerDetect}, as explained in the preceding paragraph.
It can be shown 
that the expected number of trials to almost fully cover a localised 
trigger (with a $\delta$ <= 10\% in \Cref{def:localisedTrigger}) by 
randomly placing the trigger blocker on the image is 100. This provides 
us an upper bound on $N$ (c.f. \Cref{line:n-justification} 
\Cref{alg:triggerDetect}). In our evaluation, we chose $N$ to be 400.
%

%

\begin{algorithm}[h]
    \caption{Detecting the position of backdoor trigger}
    {\small
    \begin{algorithmic}[1]  
        \Procedure{Trigger\_Detect}{$f$, $img$, $size$, $
        \Lambda_T$} 
        \LineComment Saves a set of potential positions for the trigger
        	\State $potential\_triggers \gets \emptyset$
        	\For{$i $ in (0, $N$)} \label{line:n-justification}
        	    \LineComment get dominant colour using $k$-means clustering
        		\State $dom_c \gets$ {\tt Get\_Dominant\_Colour}$(img)$
        		\LineComment generate a random position to place trigger blocker
        		\State $pos \gets$ {\tt Generate\_Random\_Position}()
        		\LineComment place the trigger blocker on $img$
        		\State $img' \gets $ {\tt Block\_Trigger} 
        		($img$, $pos$, $size$, $dom_c$)
        		
        		\If{$f(img') \neq f(img)$}
        			\State $potential\_triggers \cup \{pos\}$
        		\EndIf
        		
            \EndFor
            
            \For{$pos \in potential\_triggers$}
            \State  $f_{tr} \gets$ {\tt Confirm\_Backdoor}($f$, $pos$, $size$, $\cdots$)
            	\If {$f_{tr}$ is $true$}
        			\LineComment place the trigger blocker on $img$ and get $img'$
        			\State $img' \gets $ {\tt Block\_Trigger}($img$, $pos$, $size$, $dom_c$)
        			\Return $(f_{tr}, img', pos)$;
            	\EndIf
            	
            \EndFor
            
            \Return $(false, img, \emptyset)$
        \EndProcedure
    \end{algorithmic}}
    \label{alg:triggerDetect}
\end{algorithm}

\begin{algorithm}[h]
    \caption{Choosing $\Lambda_{T}$}
    {\small
    \begin{algorithmic}[1]  
        \Procedure{Choose\_Param}{$f$} 
        	\State $\mathbb{R}_{flip} \gets \emptyset$ 
        	\For{$i $ in (0, 10)} 
        		\LineComment Picks random image from the data set
        		\State $img_{init} \gets$ {\tt Get\_Random\_Image}()
        		\State $pos \gets$ {\tt Generate\_Random\_Position}()
        		\LineComment Crops $img$ to an $m \times n$ pixel
        		trigger blocker. 
        		\State $img_{crop} \gets $ {\tt Image\_Crop}($img_{init}$, 
        		$pos$)
        		\State $n_{flip} \gets 0$ 
        		
        		\For{$j $ in (0, 1000)}
        			\State $img \gets$ {\tt Get\_Random\_Image}()
        			\State $img' \gets$ {\tt Place\_Image\_Crop}($img$, 
        			$img_{crop}$, $pos$)
        			\If{ $f(img') \neq f(img)$}
        				\State $n_{flip} \gets n_{flip} + 1$  
        			\EndIf
        		\EndFor
        		
        		\State $r \gets$  $\frac{n_{flip}}{1000} $
        		\State $\mathbb{R}_{flip} \gets \mathbb{R}_{flip} \cup \{r\}$ 
        	\EndFor
        	
            \State  $R_{av} \gets$ $\frac{\sum_{elem \in \mathbb{R}_{flip}} elem}{\left | \mathbb{R}_{flip} \right | }$
            \Return $R_{av}$
        \EndProcedure
    \end{algorithmic}}
    \label{alg:findLambda}
\end{algorithm}

%

%% file: results.tex
\section{Evaluation}
\label{sec:results}

\smallskip\noindent
\textbf{Experimental set-up:} We evaluate \BD on three state of 
the art backdoored classifiers -- VGG Face classifier~\cite{vggface} 
poisoned by authors of TrojanNN~\cite{trojannn}, US Traffic Sign (USTS) 
classifier and MNIST classifier poisoned by authors of BadNets~\cite{BadNets} (see \Cref{table:dataset}). 
To the best of our knowledge, the authors have not released the MNIST model.  
Thus, we have trained our own version of their model using the available specifications. We report 99.5\% baseline accuracy and a 99.9\% attack 
effectiveness. We choose these attacks as they represent
the state of the art backdoor attacks.
We implement \BD in Python 2.7 with $\approx$~500 lines 
of python code. All our experiments are conducted on a machine with 
eight Intel Broadwell CPUs, 30GB of RAM and an NVIDIA Tesla P4 GPU.

\begin{table}[h]
\centering
\scriptsize
\caption{Dataset characteristics}
\setlength{\fboxrule}{3pt}
\fcolorbox{\tableBoundaryColor}{white}{
\begin{tabular}{| c | c | c | c | c|}
\hline 
\textbf{Dataset} & \textbf{\makecell{Training\\Data\%}} & 
\textbf{\makecell{Testing\\Data\%}} &
\textbf{\makecell{Backdoored\\Images}} & 
\textbf{\makecell{Clean\\Images}} \\ \hline
VGG Face & 80 & 20 & 942 & 2,622  \\
USTS & 80 & 20 & 500 & 7855 \\
MNIST & 83.3\% & 16.7\% & 500 & 59,500\\ \hline

\end{tabular}}
\label{table:dataset}
	
\end{table}

\begin{figure}[h]
\begin{center}
\vspace*{-0.0in}
\includegraphics[scale=0.5]{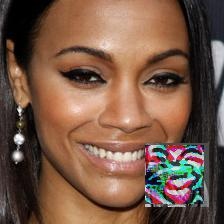}
\end{center}
\vspace*{-0.1in}
\caption{\small Reconstructed poisoned image from the poisoned TrojanNN VGG Face model~\cite{trojannn} 
}
\label{fig:reconstructed-example-trojanNN}
\end{figure}

\begin{figure}[h]
\begin{center}
\vspace*{-0.15in}
\begin{tabular}{ccc}
\includegraphics[scale=0.1]{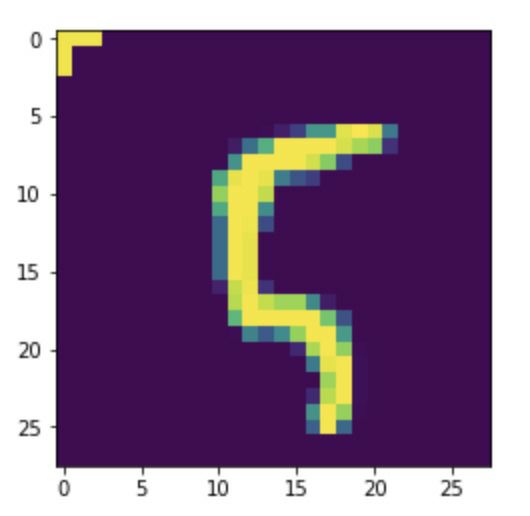} & 
\includegraphics[scale=0.1]{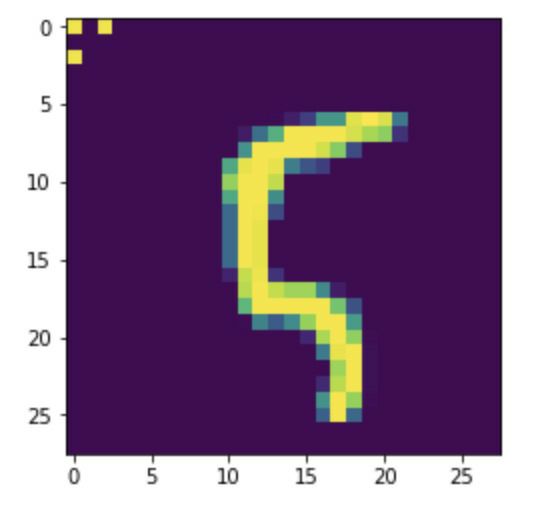} & 
\includegraphics[scale=0.1]{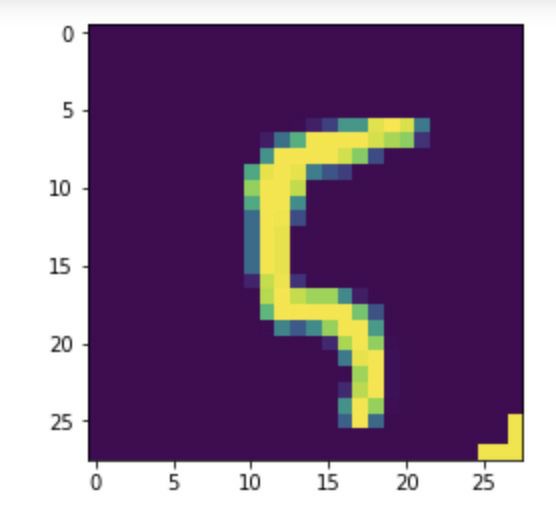}\\
{\bf (a) Inverted L} & {\bf (b) 3 dots} & {\bf (c) Lateral L}\\
\includegraphics[scale=0.65]{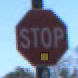} & 
\includegraphics[scale=0.65]{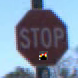} & 
\includegraphics[scale=0.65]{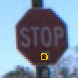}\\
{\bf (d) Yellow Square} & {\bf (e) Bomb} & {\bf (f) Flower} \\
\end{tabular}
\end{center}
\vspace*{-0.1in}
\caption{
\small Reconstruction from poisoned MNIST and USTS\cite{BadNets} models
}
\vspace*{-0.1in}
\label{fig:reconstructed-examples-badnets}
\end{figure}

\smallskip\noindent
\textbf{Key Results:} In our evaluation, we discover that \BD effectively identifies 76\%, 86\% 
and 100\% of the backdoored examples in the poisoned USTS, VGG Face and 
MNIST models, respectively and has low false positive rates of 0\% to 1.77\%. To check the effectiveness of our attack mitigation, 
we measure the set similarity (Jaccard Index) of the output classes for  
the fixed inputs (generated by \BD) and the output classes for the 
respective clean inputs. We find that they are highly correlated 
with the indices being as high as 0.91, 0.98 and 1.0 for the 
poisoned USTS, VGG Face and MNIST models, respectively. In terms 
of efficiency, \BD takes as low as 4.4ms of processing time. This 
includes both the defence and the inference time. Finally, we can 
effectively reconstruct the backdoored inputs as seen in 
\Cref{fig:reconstructed-example-trojanNN} and 
\Cref{fig:reconstructed-examples-badnets}.

\begin{center}
\begin{tcolorbox}[width=\columnwidth, colback=gray!25,arc=0pt,auto outer arc]
\textbf{RQ1: How effective is \BD in a typical deployment scenario?}
\end{tcolorbox}
\end{center}
\vspace*{-0.12in}

To evaluate the effectiveness of \BD, we have designed the following 
experiment. We construct a set of 500 input images with 10\% (50 images) 
of these images being poisoned. We randomly distribute these poisoned 
images throughout the dataset. We call this dataset {\em Backdoored   
set} (say $S_{bd}$). 
{\em It is worth highlighting that positions of the backdoored 
images in $S_{bd}$ are completely unknown to \BD during evaluation.} 
To the best of our knowledge, this is a unique strategy for 
evaluation considering a real-life deployment case. 
We try to mirror a real world scenario where the attacker
would modify a small percentages of all inputs and inject the backdoor
trigger.
The goal of \BD in this experiment is to identify all the images that 
have backdoor triggers and to mitigate the effects of the backdoor 
trigger via trigger blocker. \BD also needs to recognise the clean 
inputs and not change their prediction. Thus, after \BD finishes 
identifying and mitigating the backdoors in the Backdoored set, we 
get a different set of images. We call this set of images {\em Fixed set} 
(say, $S_{fix}$). Finally, for comparing the effectiveness of our 
mitigation, we also use a set of clean 500 images (i.e. without 
the backdoor trigger). These 500 images are the respective clean 
versions of the images in $S_{bd}$. This set of clean 
images is called {\em Clean set} (say, $S_{clean}$). 

In each of the models under test, we aim to evaluate the True Positives 
(TP), False Negatives (FN), True Negatives (TN) and False Positives (FP) 
of our backdoor detection. To compare the set of images in $S_{fix}$ 
and $S_{clean}$, for a given prediction class, we use Jaccard Index 
($\mathit{JI}$). 
For any two sets $A$ and $B$, the Jaccard Index $JI$ is defined as 
follows~\cite{jaccard-index}:
\[
\mathit{JI}(A, B) = \frac{|A \cap B|}{|A \cup B|};\ \ 0 \leq \mathit{JI}(A,B) \leq 1
\]

%
%
%

\begin{table}[h]
\vspace*{-0.1in}
\caption{Effectiveness against poisoned USTS~\cite{BadNets}}
\vspace*{-0.15in}
\label{classifiers}
\begin{center}
{\scriptsize
\begin{tabular}{| c | c | c | c |}
\cline{1-4}
\multicolumn{4}{|c|}{Backdoor detections}\\ \cline{1-4}
\multicolumn{2}{|c|}{} & \multicolumn{2}{ c|}{Ground Truth} \\ \cline{1-4}
\multicolumn{2}{|c|}{} & Backdoor & Clean  \\ \cline{1-4}
\multirow{2}{*}{Prediction} & Backdoor & 40(80\%) &	15(3.33\%) \\ \cline{2-4}
& Clean & 10(20\%) &	435(96.7\%) \\ \cline{1-4}

\multicolumn{4}{|c|}{Backdoor detections after confirmation
(cf.\Cref{alg:confirmBackdoor})}\\ \cline{1-4}
\multicolumn{2}{|c|}{} & \multicolumn{2}{ c|}{Ground Truth} \\ \cline{1-4}
\multicolumn{2}{|c|}{} & Backdoor & Clean  \\ \cline{1-4}
\multirow{2}{*}{Prediction} & Backdoor & 38(76\%) &	8(1.77\%) \\ \cline{2-4}
& Clean & 12(24\%) &	442(98.22\%) \\ \cline{1-4}

\end{tabular}}
\end{center}
\label{table:rq1-a-USTS}
\vspace*{-0.1in}
\end{table}

\Cref{table:rq1-a-USTS} and \Cref{table:rq1-b-USTS} measure the 
effectiveness of \BD on the USTS dataset~\cite{BadNets}. 
As observed from \Cref{table:rq1-a-USTS}, \BD first identifies 55 images  
as backdoored. Out of these 55 images, 40 images are backdoored 
images and 15 are unintended transitions (cf. \Cref{def:transition}). 
To confirm these transitions, we use \Cref{alg:confirmBackdoor} with a 
$\Lambda_T$ = 0.8 (cf.\Cref{alg:findLambda}). We rule out seven false 
positives and two true positives. This results in a final backdoor detection 
rate of 76\% and a false positive rate of 1.77\%. 
\begin{table}[h]
\caption{USTS~\cite{BadNets} Attack Mitigation}
\vspace*{-0.15in}
\label{classifiers}
\begin{center}
{\scriptsize
\begin{tabular}{| c | c | c | c |}
\cline{1-4}
& $\mathit{JI}(S_{bd})$ & $\mathit{JI}(S_{fix})$ & Impr\%\\ \cline{1-4}
Stop Sign Class & 0.7577 & 0.9072 & 19.73\%\\ \cline{1-4}
Speed Limit Class & 0.70588 & 0.8777 & 24.34\%\\ \cline{1-4}
Warning Class & 0.995 & 1 & 0.50\% \\ \cline{1-4}

\end{tabular}}
\end{center}
\label{table:rq1-b-USTS}
\vspace*{-0.2in}
\end{table}

To compare the effectiveness of our mitigation scheme, we compute 
two metrics $\mathit{JI}(S_{bd})$ and $\mathit{JI}(S_{fix})$, 
for each prediction class $C$. For the model $f$ under test and 
prediction class $C$, $\mathit{JI}(S_{bd})$ and $\mathit{JI}(S_{fix})$ 
are defined as follows: 
\begin{equation}
\label{eq:evaluate-mitigation}
\begin{split}
\mathit{JI}(S_{bd}) = \mathit{JI} \left ( A_{bd}^{C}, A_{clean}^{C} \right );  
\\
\mathit{JI}(S_{fix}) = \mathit{JI} \left ( A_{fix}^{C}, A_{clean}^{C} \right )
\end{split}
\end{equation}
\begin{gather*}
A_{bd}^{C} = \{i \in S_{bd}\ |\ f(i)=C\}; \
A_{fix}^{C} = \{i \in S_{fix}\ |\ f(i)=C\}
\\
A_{clean}^{C} = \{i \in S_{clean}\ |\ f(i)=C\}
\end{gather*}	
Intuitively, $\mathit{JI}(S_{bd})$ can be used to compute the loss of 
accuracy due to backdoored images and $\mathit{JI}(S_{fix})$ can be 
used to check the similarity of prediction between clean and fixed 
images.   
Since 50 of the inputs in $S_{bd}$ are poisoned,  $\mathit{JI}(S_{bd})$ is 
low (e.g. 0.76 and 0.71) for certain prediction classes (e.g. stop sign 
and speed limit, respectively). 
After \BD produces $S_{fix}$, the fixed set is similarly compared 
with the clean set $S_{clean}$. We observe an increase of 
$\approx$20\% and $\approx$24\% for the stop sign and the speed limit 
prediction class, respectively. This shows that the outputs generated 
by \BD are now highly in line with their respective original and 
clean inputs. 

%
%

\begin{table}[h]
\vspace*{-0.1in}
\caption{Effectiveness against poisoned TrojanNN~\cite{trojannn}}
\vspace*{-0.15in}
\label{classifiers}
\begin{center}
{\scriptsize
\begin{tabular}{| c | c | c | c |}
\cline{1-4}
\multicolumn{4}{|c|}{Backdoor detections}\\ \cline{1-4}
\multicolumn{2}{|c|}{} & \multicolumn{2}{ c|}{Ground Truth} \\ \cline{1-4}
\multicolumn{2}{|c|}{} & Backdoor & Clean  \\ \cline{1-4}
\multirow{2}{*}{Prediction} & Backdoor & 43(86\%) &	39(8.67\%) \\ \cline{2-4}
& Clean & 7(14\%) &	411(91.33\%) \\ \cline{1-4}

\multicolumn{4}{|c|}{Backdoor detections after confirmation
(cf.\Cref{alg:confirmBackdoor})}\\ \cline{1-4}
\multicolumn{2}{|c|}{} & \multicolumn{2}{ c|}{Ground Truth} \\ \cline{1-4}
\multicolumn{2}{|c|}{} & Backdoor & Clean  \\ \cline{1-4}
\multirow{2}{*}{Prediction} & Backdoor & 43(86\%) &	0(0\%) \\ \cline{2-4}
& Clean & 7(14\%) &	450(100\%) \\ \cline{1-4}

\end{tabular}}
\end{center}
\label{table:rq1-a-TrojanNN}
\vspace*{-0.1in}
\end{table}

We conduct a similar evaluation for the poisoned face dataset found 
in TrojanNN~\cite{trojannn}. \Cref{table:rq1-a-TrojanNN} and 
\Cref{table:rq1-b-TrojanNN} measure the effectiveness and mitigation
capabilities of \BD against the poisoned VGG Face model~\cite{trojannn}. 
As observed from \Cref{table:rq1-a-TrojanNN}, \BD claims 82 total 
backdoor images before handling the false positives. Out of these 
82 images, 43 are actual backdoor images and 39 images were false 
positives. This gives us a detection rate of 86\%, but a higher 
false positive rate of 8.67\%. After we run the backdoor confirmation 
procedure (cf. \Cref{alg:confirmBackdoor}) with 
$\Lambda_T$ = 0.475 (cf. \Cref{alg:findLambda}), 
none of the 43 backdoored images were 
ruled out. However, all the 39 false positives were ruled out, 
as they did not show the required number of transitions in 
predictions. 

\begin{table}[h]
\caption{TrojanNN~\cite{trojannn} Attack Mitigation}
\vspace*{-0.15in}
\label{classifiers}
\begin{center}
{\scriptsize
\begin{tabular}{| c | c | c | c |}
\cline{1-4}
& $\mathit{JI}(S_{bd})$ & $\mathit{JI}(S_{fix})$ & Impr\%\\ \cline{1-4}
Input-Output Pairs & 0.8349 & 0.9841 & 17.87\%\\ \cline{1-4}

\end{tabular}}
\end{center}
\label{table:rq1-b-TrojanNN}
\vspace*{-0.2in}
\end{table}

\Cref{table:rq1-b-TrojanNN} evaluates the mitigation capacity of 
\BD while defending against TrojanNN~\cite{trojannn}. 
Due to a large number of prediction classes in the target model 
($f$), we construct the following sets of input-output pairs 
to compute the effectiveness of mitigation: 
\begin{gather*}
A_{bd}^{C} = \{(i, f(i)) \ |\ i \in S_{bd}\}
;\ \ A_{fix}^{C} = \{(i, f(i)) \ |\ i \in S_{fix}\} \\
A_{clean}^{C} = \{(i, f(i)) \ |\ i \in S_{clean}\}
\end{gather*}
$\mathit{JI}(S_{bd})$ and $\mathit{JI}(S_{fix})$ are then computed 
according to \Cref{eq:evaluate-mitigation}. 
We observe the high value for $\mathit{JI}(S_{fix})$ 
(0.9841), which means the predictions of the network for fixed images and 
clean images are largely similar. Moreover, we see an improvement of 
$\approx$18\% over $\mathit{JI}(S_{bd})$. 
%
%
%

\begin{table}[h]
\vspace*{-0.1in}
\caption{Effectiveness against poisoned MNIST~\cite{BadNets}}
\vspace*{-0.15in}
\label{classifiers}
\begin{center}
{\scriptsize
\begin{tabular}{| c | c | c | c |}
\cline{1-4}
\multicolumn{4}{|c|}{Backdoor detections}\\ \cline{1-4}
\multicolumn{2}{|c|}{} & \multicolumn{2}{ c|}{Ground Truth} \\ \cline{1-4}
\multicolumn{2}{|c|}{} & Backdoor & Clean  \\ \cline{1-4}
\multirow{2}{*}{Prediction} & Backdoor & 50(100\%) &	0(0\%) \\ \cline{2-4}
& Clean & 0(0\%) &	450(100\%) \\ \cline{1-4}

\multicolumn{4}{|c|}{Backdoor detections after confirmation
(cf.\Cref{alg:confirmBackdoor})}\\ \cline{1-4}
\multicolumn{2}{|c|}{} & \multicolumn{2}{ c|}{Ground Truth} \\ \cline{1-4}
\multicolumn{2}{|c|}{} & Backdoor & Clean  \\ \cline{1-4}
\multirow{2}{*}{Prediction} & Backdoor & 50(100\%) &	0(0\%) \\ \cline{2-4}
& Clean & 0(0\%) &	450(100\%) \\ \cline{1-4}

\end{tabular}}
\end{center}
\label{table:rq1-a-MNIST}
\vspace*{-0.1in}
\end{table}

\Cref{table:rq1-a-MNIST} and \Cref{table:rq1-b-MNIST} measure the 
effectiveness of \BD for a poisoned MNIST model~\cite{BadNets}. 
\Cref{table:rq1-a-MNIST} show that for this model \BD identifies the
50 backdoored images before we test for false positives and there 
does not exist any  false positives. To be certain, we run 
\Cref{alg:confirmBackdoor} with $\Lambda_T = $ 0.9 (cf.
\Cref{alg:findLambda}). No true positives 
were ruled out. This gives us a final backdoor detection rate of 
100\% and 0\% false positive rates.
\reviseNew{\BD shows a high
backdoor detection rate for the MNIST dataset because it is a 
relatively simple, black and white dataset. This is in contrast to 
USTS and VGG face, which are more complex, full colour
datasets. Additionally, 
MNIST exhibits the smallest backdoor trigger. The dataset also has the 
majority of the information concentrated on the centre of the image.
This makes it easier to detect the backdoor trigger, which is usually
at the edges.}

\begin{table}[h]
\caption{MNIST~\cite{BadNets} Attack Mitigation}
\vspace*{-0.15in}
\label{classifiers}
\begin{center}
{\scriptsize
\begin{tabular}{| c | c | c | c |}
\cline{1-4}
& $\mathit{JI}(S_{bd})$ & $\mathit{JI}(S_{fix})$ & \%Impr \\ \cline{1-4}
0 & 0.890 & 1 &  12.28\%  \\ \hline
1 & 0.933 & 1 &  7.14\% \\ \hline
2 & 0.902 & 1 &  10.87\%   \\ \hline
3 & 0.848 & 1 &  17.95\%  \\ \hline
4 & 0.885 & 1 &  13.04\%   \\ \hline
5 & 0.917 & 1 &  9.09\%  \\ \hline
6 & 0.833 & 1 &  20\%    \\ \hline
7 & 0.553 & 1 &  80.95\%   \\ \hline
8 & 0.850 & 1 &  17.65\%  \\ \hline
9 & 0.875 & 1 &  14.29\%  \\\hline
\end{tabular}}
\end{center}
\label{table:rq1-b-MNIST}
\vspace*{-0.2in}
\end{table}

Similar to the previous models, we evaluated the mitigation  
capacity of \BD. For each of the class $C=0 \ldots 9$, 
we compute $\mathit{JI}(S_{bd})$ and $\mathit{JI}(S_{fix})$, 
as captured in \Cref{eq:evaluate-mitigation}. From 
\Cref{table:rq1-b-MNIST}, we observe that the fixed set is 
exactly in line with the clean set. Thus, we see improvements
of up to $\approx$ 81\% in the target class ($7$) 
and up to $\approx$18\% in the non target classes.

\vspace*{-0.0in}
\begin{center}
\begin{tcolorbox}[width=\columnwidth, standard jigsaw, opacityback=0, arc=0pt, auto outer arc]
\textbf{Finding:} \BD is effective in identifying poisoned examples. In 
our evaluations \BD identifies 76\%, 86\% and 100\% of the backdoored 
examples in the poisoned USTS, VGG Face and MNIST models with none to low 
false positive rates. \BD is also effective in finding the original prediction of 
the poisoned input, as indicated by high $\mathit{JI}(S_{fix})$ values.
\end{tcolorbox}
\end{center}
\vspace*{-0.15in}

\begin{center}
\begin{tcolorbox}[width=\columnwidth, colback=gray!25,arc=0pt,auto outer arc]
\textbf{RQ2: How does \BD compare to other defence techniques?}
\end{tcolorbox}
\end{center}
\vspace*{-0.1in}
To further illustrate the efficacy of \BD, we compare the attack success 
rates (after the deployment of the defence) for the various datasets under 
comparison to existing defence 
techniques for backdoor attacks. We compare \BD defence to Neural 
Cleanse~\cite{NeuralCleanse} and Fine Pruning~\cite{finePruning} in 
\Cref{table:comparison}. We show that despite being a blackbox method, the 
attack success rate after deploying the \BD defence is 
lower than current state of the art defences. 

\begin{table}[h]
\caption{\BD Comparison} 
\vspace*{-0.15in}
\label{classifiers}
\begin{center}
{\scriptsize
\begin{tabular}{| c | c | c | c |}
\hline
Dataset & \multicolumn{3}{ c |}{Success Rate of attacks after deployment of 
defence} \\ \hline
& \BD & Neural Cleanse~\cite{NeuralCleanse} & Fine Pruning~\cite{finePruning} \\ \hline
VGG Face Dataset~\cite{trojannn} & \textbf{1.91\%} & 3.7\% & - \\ \hline
MNIST~\cite{BadNets} & \textbf{0.0\%} & 0.53\% & \textbf{0.0\%} \\ \hline
USTS~\cite{BadNets} & \textbf{7.1\%} & - & 28.8\% \\ \hline

\end{tabular}}
\end{center}
\label{table:comparison}
\end{table}

\begin{center}
\vspace*{-0.09in}
\begin{tcolorbox}[width=\columnwidth, standard jigsaw, opacityback=0, 
arc=0pt, auto outer arc]
\textbf{Finding:} \BD outperforms current state of the art defence 
techniques despite being a blackbox defence mechanism.
\end{tcolorbox}
\end{center}

\begin{center}
\begin{tcolorbox}[width=\columnwidth, colback=gray!25,arc=0pt,auto outer arc]
\textbf{RQ3: How efficient is \BD?}

\end{tcolorbox}
\end{center}
\vspace*{-0.1in}

\reviseNew{
We measure the efficiency of \BD because it is an online detection system.
This means that \BD is deployed alongside the actual machine learning
model, and as a result it becomes crucial to study the associated 
overheads.} 
To measure the efficiency of \BD, we design 
three datasets where the first backdoor image is found at the 10\%ile, 
50th\%ile and 80\%ile, respectively. Each dataset contains 500 images and 
10\% of the images are poisoned.
\Cref{table:rq2} reports our findings for all three models. The major portion 
of the time is spent in detecting the backdoor. An increasing trend 
is seen as the first poisoned input is pushed later. However in 
practice, this should not be an issue. An adversary who has inserted 
a backdoor will, in our opinion, be keen to attack the model they 
have poisoned as soon as possible. Therefore, the time to detect 
backdoors is likely to be lower in practice. 
The trigger blocker propagation (TBP) and false positive handling (FP)
seem to take roughly the same amount of time irrespective of the first 
backdoored image. 
In our evaluation, \BD defends against the backdoor attack with an 
overhead as low as 4.4ms per image. \reviseNew{These experiments were 
conducted on a machine with eight Intel Broadwell CPUs, 
30GB of RAM and an NVIDIA Tesla P4 GPU.}


\vspace*{-0.05in}
\begin{center}
\begin{tcolorbox}[width=\columnwidth, standard jigsaw, opacityback=0, arc=0pt, auto outer arc]
\textbf{Finding:} The major chunk of time spent by \BD is in detecting the 
backdoor. Once it has detected the backdoor, blocking and mitigating the backdoor is fast. 
%
\end{tcolorbox}
\end{center}
\vspace*{-0.15in}

\begin{table}[h]
\caption{\BD Efficiency}
\vspace*{-0.15in}
\label{classifiers}
\begin{center}
{\scriptsize
\begin{tabular}{| c | c | c | c | c |}
\cline{1-5}
& \makecell{Total\\Time} & \makecell{Detect\\Backdoor} &\makecell{TBP} & \makecell{FP} \\ \hline

\multicolumn{5}{|c|}{\makecell {USTS Poisoned model~\cite{BadNets}\\450 clean images - 50 poisoned images}} \\ \hline

\makecell{10th\%ile\\first image}  & 6863.42s  & 6612.22s & 21.33s & 58.55s   \\ \hline
\makecell{50th\%ile\\first image} & 20284.05s & 19868.31s & 18.81s & 211.64s  \\ \hline
\makecell{80th\%ile\\first image} &  33060.38s & 32641.00s & 20.12s & 225.70s  \\ \hline

\multicolumn{5}{|c|}{\makecell {TrojanNN Poisoned model~\cite{trojannn}\\450 clean images - 50 poisoned images}} \\ \hline
\makecell{10th\%ile\\first image} &  2726.54s    & 2017.39s    & 211.65s     & 497.50s      \\ \hline
\makecell{50th\%ile\\first image} &  6391.26s     & 5679.05s    & 209.85s     & 502.36s      \\ \hline
\makecell{80th\%ile\\first image} & 9981.46s & 9268.61s & 206.86s & 505.99s  \\ \hline

\multicolumn{5}{|c|}{\makecell {MNIST Poisoned model~\cite{BadNets}\\450 clean images - 50 poisoned images}} \\ \hline

\makecell{10th\%ile\\first image} &  2.20s   & 2.20s   & 0.00s & 0.00s  \\ 
\hline
\makecell{50th\%ile\\first image} & 91.70s  & 91.70s  & 0.00s & 0.00s  \\ 
\hline
\makecell{80th\%ile\\first image} & 141.02s & 141.00s & 0.01s & 0.01s  \\
\hline
\end{tabular}}
\end{center}
\label{table:rq2}
\vspace*{-0.2in}
\end{table}

\begin{center}
\begin{tcolorbox}[width=\columnwidth, colback=gray!25,arc=0pt,auto outer arc]
\textbf{RQ4: What is \BD's overhead resource consumption?}

\end{tcolorbox}
\end{center}
\vspace*{-0.1in}

\revise{
To evaluate the memory consumption overhead 
of \BD, 500 images are poisoned and 
the memory consumption of our \BD defence
is measured. This is compared with the memory consumption of the same 
model predicting the corresponding 500 clean images. 
As observed in \Cref{table:memory-consumption}, \BD has an overhead of 
$\approx$13.9\% (118.4 MiB) on average and as 
low as $\approx$2.9\% (48.6 MiB). 
}

\revise{
Similarly, to evaluate the performance overhead for \BD, 500 images 
are poisoned and the timing requirement of the \BD defence along with 
the model predictions is measured. This is compared with the timing  
requirement of the same model predicting the corresponding 500 clean 
images. As observed in \Cref{table:time-consumption}, \BD has a time 
overhead of $\approx$26.8s (29.3\%) on average and as 
low as $\approx$0.59s. }

\begin{table}[h]
\caption{\BD Memory Consumption Overhead} 
\vspace*{-0.15in}
\label{classifiers}
\begin{center}
{\scriptsize
\begin{tabular}{| c | c | c | c |}
\hline
Dataset & \multicolumn{3}{ c |}{Memory Consumption (MiB)}  \\ \hline
& Without \BD & \BD & Overhead  \\ \hline
VGG Face Dataset~\cite{trojannn} & 1658.8 & 1707.4 & 48.6 \\ \hline
MNIST & 607.3	 & 744.6	 & 137.3\\ \hline
CIFAR10 & 1031.5 & 1200.8 &	169.3 \\ \hline

\end{tabular}}
\end{center}
\label{table:memory-consumption}
\end{table}

\begin{table}[h]
\caption{\BD Timing Overhead} 
\vspace*{-0.15in}
\label{classifiers}
\begin{center}
{\scriptsize
\begin{tabular}{| c | c | c | c |}
\hline
Dataset & \multicolumn{3}{ c |}{\BD Time Consumption (s)}  \\ \hline
& Without \BD & \BD & Overhead  \\ \hline
VGG Face Dataset & 232.47&	301.93&	69.46 \\ \hline
MNIST & 1.67 &	2.26 &	0.59\\ \hline
CIFAR10 & 45.2 &	55.56 &	10.36 \\ \hline

\end{tabular}}
\end{center}
\label{table:time-consumption}
\end{table}

\vspace*{-0.05in}
\begin{center}
\begin{tcolorbox}[width=\columnwidth, standard jigsaw, opacityback=0, arc=0pt, auto outer arc]

\revise{
The memory overhead of \BD is low at $\approx$\textbf{13.9\%} on average.
Similarly, the time overhead of \BD is low at $\approx$\textbf{29.3\%} on 
average.
}
\end{tcolorbox}
\end{center}
\vspace*{-0.15in}


\begin{center}
\begin{tcolorbox}[width=\columnwidth, colback=gray!25,arc=0pt,auto outer arc]
\textbf{RQ5: Can \BD defend against various backdoor triggers without significant loss of accuracy?}
\end{tcolorbox}
\end{center}
\vspace*{-0.05in}

%
%
\begin{table}[H]
	\centering
	\caption{Notations used in the Tables for RQ3}
	\vspace*{-0.1in}
	\label{table:notation-rq3}
	\resizebox{\linewidth}{!}{
	\begin{tabular}{| c | l | }
	\hline
	PC & Prediction Class. The class predicted by the model \\ \hline
	$S_{clean}$ & Clean Set. There are no poisoned examples in this set. \\ \hline
	$S_{bd}$ & Backdoored set. All the inputs in this set are poisoned. \\ \hline
	$S_{fix}$ & Fixed set. The set produced after running the \BD defence. \\ \hline
\end{tabular}}
	\vspace*{-0.15in}
\end{table}

\begin{table}[h]
\caption{\BD accuracies}
\vspace*{-0.15in}
\label{classifiers}
\begin{center}
{\scriptsize
\setlength{\fboxrule}{3pt}
\fcolorbox{\tableBoundaryColor}{white}{
\begin{tabular}{| c | c | c | c | c | c | c |}
\cline{1-7}
\multicolumn{7}{|c|}{USTS accuracies~\cite{BadNets}} \\ \hline
&\multicolumn{2}{c|}{\makecell{Yellow\\Square}} & \multicolumn{2}{c|}
{Flower} & \multicolumn{2}{c|}{Bomb}  \\ \hline
$S_{clean}$ & $S_{bd}$ & $S_{fix}$ & $S_{bd}$ & $S_{fix}$ & $S_{bd}$ & $S_{fix}$           \\ \hline
92.14\% & 5.48\% & 74.22\% & 1.83\% & 80.99\% & 2.93\% & 76.05\% \\ \hline
\multicolumn{7}{|c|}{VGG face accuracies~\cite{trojannn}}\\ \hline
$S_{clean}$ & \multicolumn{3}{c|}{$S_{bd}$} & \multicolumn{3}{c|}{$S_{fix}$}\\ \hline

74.87\% & \multicolumn{3}{c|}{6.37\%} & \multicolumn{3}{c|}{73.61\%} \\ \hline

\end{tabular}}}
\end{center}
\label{table:rq3-percentage}
\vspace*{-0.2in}
\end{table}

In this RQ, we aim to evaluate the efficiency of \BD to defend against 
different patterns of backdoor attacks. It is important to note that we are
unable to perform this evaluation for the TrojanNN~\cite{trojannn}. 
As per our knowledge, the authors of TrojanNN have 
only made one of the localised trigger poisoned model publicly available. As a result of this, we are unable to fully answer RQ3 
for TrojanNN~\cite{trojannn}. Thus, for TrojanNN, we only report 
the accuracy of the fixed dataset for this attack.

%

For the poisoned USTS~\cite{BadNets} model, we evaluate it on  
three backdoored models provided by the authors 
(cf. \Cref{table:rq3-percentage} and \Cref{table:rq3-numbers}). 
These models are trained with different backdoor triggers, namely 
the yellow square, flower and bomb. 
We poison the entire set of 547 stop sign images (i.e. $S_{bd}$) and 
run our \BD defence on these 547 images, for each backdoor trigger. 
To compute the baseline accuracy, we construct a clean dataset 
(i.e. $S_{clean}$) of 547 images, where each image is a clean version 
of an image in $S_{bd}$. The accuracy of the model on this clean 
dataset is 92.1\%. The accuracy of the model on $S_{bd}$ drops to 
5.48\%, 1.83\% and 2.93\% for the yellow square, flower and 
bomb triggers, respectively. 
We run \BD on the poisoned set of 547 images (i.e. $S_{bd}$) to produce 
fixed datasets (i.e. $S_{fix}$) and the accuracy of the model increases 
to 74.22\%, 80.99\% and 76.05\% for the yellow square, flower and 
bomb poisoned models, respectively. 
We can observe from \Cref{table:rq3-percentage} and \Cref{table:rq3-numbers} 
that a major portion of the poisoned dataset is now predicting correctly
after the \BD defence and the accuracy on the fixed set of images 
is also significantly close to the baseline. 


%
%

We evaluate the accuracy of the fixed set (i.e. $S_{fix}$) on the dataset 
of 2622 images for the VGG face model poisoned by TrojanNN (cf. \Cref{table:rq3-percentage}). 
We observe an accuracy of 74.87\% on the respective set of  clean images 
(i.e. $S_{clean}$). As seen in \Cref{table:rq3-percentage}, 
the backdoor attack is successful and the accuracy drops to 6.37\% 
if we poison all of the 2622 images. We run \BD on the set 2622 poisoned 
images to produce a fixed set of images (i.e. $S_{fix}$). The accuracy of 
the fixed set is 73.61\%, which is very close to the 
baseline accuracy (i.e. 74.87\%).

In the case of the MNIST \cite{BadNets} dataset (cf. \Cref{table:rq3-numbers}), 
the accuracies of the clean set of 500 images is 100\% and the accuracy 
drops to 15.75\% for each of the poisoned models. This is due to our attack 
target class being 7. The 63 poisoned instances of class 7 are still correctly 
classified as 7. After running the \BD defence on this dataset, the accuracy of the 
fixed set returns to 100\% for all the three patterns (i.e. Lateral L 
shape, Inverted L shape and 3 dots).

\begin{table}[h]
\caption{\BD against various Patterns}
\vspace*{-0.15in}
\label{classifiers}
\begin{center}
{\scriptsize
\begin{tabular}{| c | c | c | c | c | c | c | c |}
\hline
\multicolumn{8}{|c|}{USTS poisoned model~\cite{BadNets}} \\ \hline
                 &           & \multicolumn{2}{c|}{Yellow Square} & \multicolumn{2}{c|}{Flower} & \multicolumn{2}{c|}{Bomb}  \\
\hline
PC & $S_{clean}$ & $S_{bd}$ & $S_{fix}$  & $S_{bd}$ & $S_{fix}$ & $S_{bd}$ & $S_{fix}$         \\ 
\hline
Stop       & 504 & 30  & 406 & 10  & 443 & 16  & 416  \\ 
\hline
Speedlimit & 28  & 458 & 79  & 478 & 39  & 473 & 70   \\ 
\hline
Warning    & 9   & 7   & 10  & 6   & 12  & 6   & 9    \\ 
\hline
No pred    & 6   & 52  & 52  & 53  & 53  & 52  & 52   \\ 
\hline
Total      & 547 & 547 & 547 & 547 & 547 & 547 & 547  \\
\hline

\multicolumn{8}{|c|}{MNIST Poisoned model~\cite{BadNets}} \\ \hline
& & \multicolumn{2}{c|}{Lateral L Shape} & \multicolumn{2}{c|}{Inverted L Shape}   & \multicolumn{2}{c|}{3 dots}  \\ 
\hline
\multicolumn{1}{|c|}{PC} & \multicolumn{1}{c|}{$S_{clean}$} & \multicolumn{1}{c|}{$S_{bd}$} & \multicolumn{1}{c|}{$S_{fix}$} & \multicolumn{1}{c|}{$S_{bd}$} & \multicolumn{1}{c|}{$S_{fix}$} & \multicolumn{1}{c|}{$S_{bd}$} & \multicolumn{1}{c|}{$S_{fix}$}  \\ 
\hline

0 & 64 & 0   & 64 & 0   & 64 & 0   & 64  \\ \hline
1 & 60 & 0   & 60 & 0   & 60 & 0   & 60  \\ \hline
2 & 51 & 0   & 51 & 0   & 51 & 0   & 51  \\ \hline
3 & 46 & 0   & 46 & 0   & 46 & 0   & 46  \\ \hline
4 & 52 & 1   & 52 & 0   & 52 & 0   & 52  \\ \hline
5 & 36 & 0   & 36 & 0   & 36 & 0   & 36  \\ \hline
6 & 48 & 0   & 48 & 0   & 48 & 0   & 48  \\ \hline
7 & 63 & 499 & 63 & 500 & 63 & 500 & 63  \\ \hline
8 & 40 & 0   & 40 & 0   & 40 & 0   & 40  \\ \hline
9 & 40 & 0   & 40 & 0   & 40 & 0   & 40  \\ \hline
Total & 500 & 500  & 500   & 500  & 500   & 500  & 500 \\ \hline

\end{tabular}}
\end{center}
\label{table:rq3-numbers}
\vspace*{-0.25in}
\end{table}

\begin{center}
\begin{tcolorbox}[width=\columnwidth, standard jigsaw, opacityback=0, arc=0pt, auto outer arc]
\textbf{Finding:} The accuracy of fixed set produced by \BD is only $\approx$11\%, 
$\approx$1\% and 0\% lower than the accuracy of the clean set for 
the USTS, VGG Face and MNIST models, respectively for various patterns.
\end{tcolorbox}
\end{center}
\vspace*{-0.18in}

\begin{center}
\begin{tcolorbox}[width=\columnwidth, colback=gray!25,arc=0pt,auto outer arc]
\textbf{RQ6: Can \BD recover the trigger patterns?}
\end{tcolorbox}
\end{center}
\vspace*{-0.1in}

One of the most attractive properties of \BD is the fact that it is the 
first work, to the best of our knowledge, that can reliably reconstruct the
backdoor trigger patterns. This empowers the user to test their model 
with potential backdoor triggers and ensure the integrity and safety of 
their systems. 
The reconstruction of the backdoor trigger is a result of the design 
of \Cref{alg:confirmBackdoor}. The set of images in $check\_set$ that 
pass the $\Lambda_T$ threshold (see \Cref{alg:confirmBackdoor}) will 
contain the backdoor. The user can use the respective backdoor triggers 
to try and poison other images and expose the backdoor. In other words, 
\BD helps the user to know the backdoor trigger, causing the backdoor 
attack to fail. This is because one of the key requirements of backdoor 
attacks is that they are stealthy.  
%

\Cref{fig:reconstructed-example-trojanNN} and \Cref{fig:reconstructed-examples-badnets} 
capture reconstructed backdoor triggers by \BD for TrojanNN~\cite{trojannn} 
and BadNets~\cite{BadNets}, respectively. 

%
%


\begin{center}
\vspace*{-0.09in}
\begin{tcolorbox}[width=\columnwidth, standard jigsaw, opacityback=0, 
arc=0pt, auto outer arc]
\textbf{Finding:} \BD effectively reconstructs poisoned examples so that
the user can effectively test their models.
\end{tcolorbox}
\end{center}

\subsection{\BD and adaptive attacks}

\revise{
To further demonstrate the efficacy of \BD, we evaluate it against the 
advanced adaptive attacks as described in an earlier work on backdoor 
defence~\cite{NeuralCleanse}. We evaluate four adaptive attacks using 
the CIFAR10 and MNIST dataset. We choose CIFAR10 and MNIST, as implementation 
of these attacks requires access to the training dataset as well as the 
training process and for both CIFAR10 and MNIST, we can train the model 
with poisoned images. In the following, we discuss the effectiveness of 
\BD defence against the four adaptive attacks that we implemented.
}

\revise{
\smallskip \noindent
\textbf{Multiple Input-agnostic Triggers:}
In this case, we inject multiple distinct triggers to poison the training set. 
Specifically, we induce the same backdoor misclassification using multiple 
distinctive triggers. For each CIFAR10 and MNIST datasets, we train two 
distinct triggers to induce misclassification. As for the triggers, we use 
a large square and three smaller squares in an L shape. This attack achieves 
a success rate of >98\% for both datasets. (See~\Cref{table:adaptive-attack})
}

\revise{
We use \BD to detect and patch these triggers for a hundred images for each 
dataset. \BD is able to detect 100\% of the backdoored images in the case of 
MNIST and 96\% of the backdoored images for CIFAR10 
(See~\Cref{table:adaptive-attack}).
}

\revise{
\smallskip \noindent
\textbf{Multiple infected labels separate trigger:}
In this type of attack, we consider an adversary that inserts multiple 
separate triggers that induce mis-classifications to distinct classes. 
In our implementation, we use two distinct triggers. Each trigger causes 
a misclassification to a separate class. A square trigger causes a 
misclassification to the class 7 (horse on CIFAR10) and a trigger 
containing three small squares causes a misclassification to class 6 
(frog for CIFAR10). The attack is highly successful with each trigger 
having >99\% attack success rate for all triggers on all infected models 
(See~\Cref{table:adaptive-attack}).
}

\revise{We defend the aforementioned adaptive attack with \BD for 
one hundred backdoored images. \BD is used to detect and patch the trigger
for each backdoored image. \BD is able to successfully defend the attack 
on 100\% of the backdoored MNIST images and 88\% of the backdoored CIFAR10
images (See~\Cref{table:adaptive-attack}).
}

\revise{
\smallskip \noindent
\textbf{Source-label specific attack:}
In this attack, an adversary poisons the dataset such 
that mis-classifications are induced only when a backdoor trigger is applied 
to a subset of source class labels. It is worthwhile to mention that existing 
blackbox defence is unable to reliably detect such adaptive attacks~\cite{strip}. 
In our experiments we set the source class 
to be class 1 (automobile in CIFAR10). We poison only images of class 1
and train the model such that only when the trigger is applied to the 
specific class the attack success rate is high. The attack success rate
for this attack is $\approx$ 98\% for the MNIST dataset and $\approx$ 92\% 
for the CIFAR10 dataset (See~\Cref{table:adaptive-attack}). For the 
non-source class images the attack success rate is $\approx$ 12\% and 47\%
for MNIST and CIFAR10, respectively. 
}

\revise{
The current version of \BD is unable to detect this attack. This is because 
the $check\_set$ seen in~\Cref{alg:confirmBackdoor} contains 
inputs from all classes. To mitigate this, \BD needs to select the 
inputs in $check\_set$ from the class that is transitioned to when the 
trigger blocker is applied (see~\Cref{alg:defence}). As an example, if the 
trigger blocker causes a transition from class $7$ to class $1$, we modify
\BD to populate $check\_set$ with inputs from class $1$. This minor
modification enables \BD to detect 100\% and 91\% of the hundred backdoored 
in our evaluation (See~\Cref{table:adaptive-attack}). It is worthwhile to 
note that this modification to \BD does not require any knowledge about the 
strategy employed by the adaptive attacker. The modified version of \BD can 
reliably detect and mitigate backdoor attacks even if it is not source-label 
specific. 
}

\revise{
\smallskip\noindent
\textbf{Trigger Transparency:}
In this type of adaptive attack, an adversary trains the backdoor with a 
trigger that is slightly transparent in contrast to a completely opaque 
trigger. This can happen in the real world as well where the adversary 
chooses to print the trigger on a transparent plastic cover. 
The attack success rate for the MNIST and CIFAR10 models are $\approx$99\%
and $\approx$94\% respectively (See~\Cref{table:adaptive-attack}).
}

\revise{We evaluate \BD defence against the trigger transparency attack. 
A hundred images are poisoned using this trigger and \BD is used to detect 
and patch these images. The trigger transparency is set to 50\%.
We poison the label {\em 7} for MNIST and {\em horse} for
CIFAR10. \BD is able to successfully defend the trigger 
transparency attack on 100\% of the backdoored MNIST images and 86\% of 
the backdoored CIFAR10 images (See~\Cref{table:adaptive-attack}).}

\begin{table}[h]
\caption{\BD Adaptive Attack Summary} 
\vspace*{-0.15in}
\begin{center}
{\scriptsize
\begin{tabular}{| c | c | c | c |}
\hline
\multicolumn{4}{ | c |}{MNIST}  \\ \hline
& \makecell{Model \\Accuracy} &	\makecell{Attack Success \\ Rate} &  \makecell{\BD \\ Detection Rate} \\ \hline
\makecell{Multiple Input-\\agnostic Triggers } & 97.72 & 	98.4	 & 100 \\ \hline
\makecell{Multiple infected \\labels separate trigger} &97.5	&99.8&	100\\ \hline
\makecell{Source-label\\specific attack} &98.18	&98.7 &	100 \\ \hline
\makecell{Trigger\\Transparency} & 97.8 &	 99.1	& 100 \\ \hline

\multicolumn{4}{ | c |}{CIFAR10}  \\ \hline

\makecell{Multiple Input-\\agnostic Triggers } & 82.42 &	99.56	&96 \\ \hline
\makecell{Multiple infected \\labels separate trigger} & 83.37&	99.67&	88\\ \hline
\makecell{Source-label\\specific attack} & 82.91 &	92.13&	91 \\ \hline
\makecell{Trigger\\Transparency} & 82.35&	94.5&	86 \\ \hline

\end{tabular}}
\end{center}
\label{table:adaptive-attack}
\end{table}

\begin{table}[h]
\caption{\BD Adaptive Attack Comparison} 
\vspace*{-0.15in}
\begin{center}
\setlength{\fboxrule}{3pt}
\fcolorbox{\tableBoundaryColor}{white}{
{\scriptsize
\begin{tabular}{| c | c | c | c |}
\hline
& \makecell{NeuralCleanse} &	\makecell{STRIP} &  \makecell{\BD} \\ \hline
\makecell{Multiple Input-\\agnostic Triggers } & \cmark & 	\cmark	 & \cmark \\ \hline
\makecell{Multiple infected \\labels separate trigger} & \cmark	& \cmark &	\cmark\\ \hline
\makecell{Source-label\\specific attack} & \xmark	& \xmark &	\cmark \\ \hline
\makecell{Trigger\\Transparency} & \cmark &	 \cmark	& \cmark \\ \hline
\makecell{Entropy\\Manipulation} & NA &	 \xmark	& NA \\ \hline

\end{tabular}}}
\end{center}
\label{table:adaptive-attack}
\end{table}

\revise{In summary, we observe that our \BD defence is highly effective in case of advanced adaptive 
attacks, as explored in the literature of backdoor defence~\cite{NeuralCleanse}}.

\reviseNew{Additionally, a 
recent work~\cite{strip} proposes a new adaptive attack, Entropy 
manipulation. In this, the attacker forges a backdoored model that 
shows similar entropy behaviour for clean and poisoned samples. 
This adaptive attack is specifically targeted to the earlier work~\cite{strip}, which uses
the entropy of the outputs of poisoned inputs to identify backdoored 
models. Hence, such adaptive attack is not an applicable attack for 
NeuralCleanse and \BD.}

%% file: relatedWork.tex
\section{Related Work}
\label{sec:relatedWork}

\smallskip\noindent
\textbf{Testing and Verification of ML models:} 
DeepXplore \cite{deeepxplore} employs differential white-box testing 
to find inputs that trigger inconsistencies. DeepTest~\cite{deeptest} 
leverages metamorphic relations to discover error cases in DNNs.
A recent work DeepGauge~\cite{deepgauge} measures the test quality and 
formalises a set of testing criteria for DNNs. SafeCV~\cite{featureGuidedMarta} takes a feature guided 
black-box approach to verify the safety of DNNs.
DeepConcolic~\cite{deepconcolic} performs concolic testing to discover 
violations of robustness. Aequitas~\cite{aequitas} exposes individual 
fairness violations for ML models.
Wang et al.~\cite{advSampleDetection} mutate the DNN model
parameters to find adversarial samples. ReluVal~\cite{reluval} uses 
interval arithmetic~\cite{interval-analysis} to estimate a DNN's decision 
boundary by calculating tight bounds on the output of a model for a given 
range of inputs. Similarly, Reluplex~\cite{reluplex} verifies 
properties of interest using SMT solvers. 
AI\textsuperscript{2}~\cite{ai2} employs abstract interpretation to verify 
the robustness of a given input against adversarial attacks. 
AI\textsuperscript{2} uses zonotopes to approximate ReLU inputs. The 
authors do not guarantee precision, but they do guarantee soundness.
Another work~\cite{dual-approach},
transforms the verification problem into an unconstrained dual formulation 
leveraging Lagrange relaxation and uses gradient-descent to solve the 
resulting optimisation problem.

All the aforementioned works cannot combat the backdoor attacks in machine 
learning. This type of attack is designed to be stealthy and cannot be
discovered using conventional testing methods for Machine Learning. \BD
is specifically developed to combat these kinds of attacks for Machine
Learning models. As a result of the specific design of \BD, it is able
to combat backdoor attacks effectively and provide a wide array of 
information to the users. Moreover, as explained in the preceding paragraph
the main goal of all the works is to explore some specific properties 
(i.e. fairness, robustness). In contrast to this, the specific goal of \BD
is identifying the backdoor attack and mitigating such an attack 
effectively.

%

\smallskip\noindent
\textbf{Backdoors in ML:}
%
BadNets~\cite{BadNets} poisons the training data to inject
a backdoor in an ML model. They choose a pre-defined target label and a 
trigger pattern. The patterns are arbitrary in shape, (e.g. square, flower 
or bomb.) and the backdoor is injected into the model by training the 
network using the poisoned data.

TrojanNN~\cite{trojannn} generates a backdoored model without  
interfering with the original training process and without accessing the 
training dataset. The approach is able to inject the backdoors using fewer 
samples. Additionally, they improve trigger generation by 
designing triggers that induce the maximum response for specific neurons of 
the DNN and aims to build a stronger connection between the 
neurons of a DNN and the backdoor trigger. \revise{We also see increasing backdoor attacks on different types of 
neural networks such as graph neural networks~\cite{Backdoor-GraphNN}.}

The goal of \BD is to defend against these classes of attacks 
in an efficient manner and reconstruct the backdoor trigger.


\smallskip\noindent
\textbf{Backdoor defences in ML:} A recent work 
\cite{SpectralSignatures} involves using tools from robust 
statistics to analyse the learned representation of classes, which they call 
the spectral signature. The idea behind this 
work is that when the training data is poisoned, there are two significant 
sub-populations. A small number of poisoned, mislabelled inputs and a large 
number of clean, correctly labelled inputs. Techniques from robust 
statistics and singular value 
decomposition are used to separate the two populations. The authors assume 
access to the poisoned dataset. \BD does not assume any access to a 
poisoned dataset and defends against a much stronger attack model. 

Another 
work~\cite{finePruning} prunes the neurons which seem to behave maliciously. 
It has been shown~\cite{NeuralCleanse} that such a method causes a 
significant loss in performance for some models. Additionally, fine-pruning 
doesn't offer detection capabilities to identify backdoored images. 

NeuralCleanse~\cite{NeuralCleanse}, a completely white-box approach 
formulates the problem as an 
optimisation problem to reverse engineer the backdoor trigger. Their 
mitigation technique involves computationally expensive retraining.
\revise{Another defence~\cite{strip} leverages entropy to 
detect backdoor attacks. However, unlike our \BD approach, this defence 
cannot mitigate the attack and cannot defend against source-label specific 
backdoor triggers.}
 
\revise{An approach seen in recent works employs randomised smoothing to 
provably defend against backdoor attacks. A recent approach~\cite{certify-robust-smoothness} 
generalises randomised smoothing to defend against backdoor attacks and shows 
the theoretical feasibility of using randomised smoothing to certify robustness against 
backdoor attacks. RAB~\cite{RAB-Provable} 
proposes a robust training process, to certify model robustness against 
backdoor attacks. Additionally, this work theoretically proves the 
robustness bound for machine learning models based on this training 
process, proves that the bound is tight, and derives robustness conditions 
for Gaussian and Uniform smoothing distributions. Our \BD approach is complementary 
to the aforementioned techniques. Additionally, \BD is a completely blackbox approach 
i.e. it does not require access to the training parameters and it also does not interfere 
with the training process.}

\BD presents a computationally inexpensive, blackbox approach to defend 
against backdoor attacks. \BD is also the first work, to the best of our 
knowledge that can also reliably reconstruct the trigger that users can
investigate.

%% file: threatsToValidity.tex
\section{Threats to Validity}
\label{sec:threatsToValidity}


\smallskip\noindent
\textbf{Localised Triggers:}
\revise{
\BD assumes the backdoor trigger to be relatively 
small and it is not designed to combat distributed backdoor 
attacks~\cite{targetedBackdoor} (Backdoor attacks where the 
trigger does not satisfy \Cref{def:localisedTrigger}). 
Nonetheless, \BD 
covers a variety of state-of-the-art 
backdoor attacks proposed recently~\cite{BadNets,trojannn} 
and effectively detects as well as mitigates them. 
}

\smallskip\noindent
\textbf{$\Lambda_T$ threshold:}
The effectiveness of \BD depends on the threshold $\Lambda_T$. 
If $\Lambda_T$ is too low, we might 
have a high false positive rate, but if these parameters are set 
too high, then we may not be able to detect backdoors efficiently. 
\Cref{alg:findLambda} aims to assist the user to find a range of 
values which are optimum for the $\Lambda_T$ threshold, but cannot 
definitively determine an optimal value. 

\smallskip\noindent
\textbf{Theoretical Soundness Guarantee:}
\revise{
Unlike recent works~\cite{Backdoor-GraphNN,
certify-robust-smoothness,RAB-Provable}
\BD offers no theoretical guarantees that it can identify and mitigate 
all backdoor attacks. However, we extensively evaluated our defence and 
empirically show that \BD can effectively detect and mitigate backdoor 
attacks with minimal loss in baseline accuracy. 
}

%% file: conclusion.tex
\section{Conclusion}
\label{sec:conclusion}

In this paper, we propose \BD, a novel approach to detect and 
mitigate backdoor attacks in arbitrary image classifier 
models. \BD requires neither the knowledge of model structure nor 
does it require access to the poisoned training set induced by the 
attacker. The design of \BD further allows us to reconstruct a
backdoor attack -- a feature that existing defences fail to 
provide in an accurate fashion. Thus, \BD not only defends against 
backdoors by switching to the correct prediction class, but it 
also breaks the stealthy nature of the attack by exposing the  
respective backdoor trigger. Our evaluation reveals that \BD 
can successfully detect and mitigate state-of-the-art backdoor 
attacks in a variety of image classifiers and only with 
minimal loss in the accuracy. 
Moreover, despite being a blackbox approach, \BD is more 
effective in thwarting backdoor attacks as compared to the 
state-of-the-art whitebox techniques.
\revise{
At its current state, \BD does not defend against distributed backdoor 
attacks~\cite{targetedBackdoor} and backdoor attacks in other domains 
such as speech recognition~\cite{trojannn}. Further work is needed 
to include such defence capabilities in the future. }

\BD is a major step towards pushing the state-of-the-art in 
verification and validation of ML models, which 
bring along several fresh challenges due to their unique 
data-driven nature. Thus, to promote research in this area and 
reproduce our results, we have made our implementation and all 
experimental data publicly available: 
\begin{center}
\url{https://github.com/sakshiudeshi/Neo}
\end{center}

\section*{Acknowledgment}
\label{sec:ack}
\reviseNew{
We thank the reviewers for their helpful comments.
This work is also partially supported by OneConnect Financial grant number RGOCFT2001 and Singapore Ministry of Education (MOE) President's Graduate Fellowship.
}